\documentclass[journal, twoside]{IEEEtran}   

\usepackage[hidelinks]{hyperref}    
\usepackage[cmex10]{amsmath}        
\usepackage{amssymb,amsfonts}       
\usepackage{graphicx}
\usepackage{subcaption}             
\usepackage{booktabs}               
\usepackage{multirow}
\usepackage{tabularx}
\usepackage{siunitx}                
\usepackage{dblfloatfix}            
\usepackage{bm,bbm}
\usepackage{orcidlink}
\usepackage[numbers,compress]{natbib}

\graphicspath{{Figures/PDF/}{Figures/PNG/}}
\usepackage{texnames} 
\usepackage{verbatim}
\usepackage{makecell}
\usepackage{algorithm}
\usepackage{algpseudocode}

\begin{document}

\title{FusionNet: Physics-Aware Representation Learning for Multi-Spectral and Thermal Data via Trainable Signal-Processing Priors}

\author{Georgios~Voulgaris\,\orcidlink{0000-0003-4597-7352}%
\thanks{Manuscript submitted to the \textit{IEEE Journal of Selected Topics in Applied Earth Observations and Remote Sensing (J-STARS)}, Special Issue on IGARSS~2025. (Corresponding author: Georgios Voulgaris.)

Georgios Voulgaris is with the University of Oxford, Oxford, United Kingdom (e-mail: georgios.voulgaris@biology.ox.ac.uk).}%
}

\markboth{IEEE Journal of Selected Topics in Applied Earth Observations and Remote Sensing,~Vol.~19,~2026}{Voulgaris: FusionNet: Multi-Spectral and Thermal Deep Learning Fusion for Cement Plant Detection}



\maketitle


\begin{abstract}
     Cement production underpins global infrastructure but contributes approximately 7$\%$ of anthropogenic CO$_2$ emissions, making accurate monitoring of production facilities essential for sustainable development. Existing remote sensing approaches rely predominantly on thermal signatures from kiln operations, which can be confounded by background heat sources and fail to capture persistent environmental alterations. This study introduces a physics‑informed methodology that exploits multi-spectral features, particularly a geological Short Wave Infrared (SWIR) ratio, to detect soil property changes induced by sustained industrial heat emissions. This work proposes FusionNet, an intermediate multi-spectral data fusion framework that integrates Thermal Infrared (TIR) and SWIR inputs through a specialised backbone, embedding differential signal processing priors within a convolutional layer, mixed pooling, and wider receptive field. Systematic ablation studies confirm that each architectural component contributes to performance gains, with DGCNN achieving a 4.1-6.8$\%$ accuracy improvement over conventional CNNs. On the SWIR ratio dataset, FusionNet attains a maximum of 90.6$\%$, outperforming state‑of‑the‑art baselines across five spectral configurations and exceeding the strongest unimodal model by 1.1$\%$. Transfer learning experiments reveal that ImageNet pretraining degrades TIR and SWIR performance, underscoring the importance of modality‑aware training for cross‑spectral applications. Overall, the results demonstrate that combining physics‑aware feature selection with principled deep learning architectures enables robust, high‑accuracy detection of cement production facilities, offering a reliable framework for industrial infrastructure monitoring.
\end{abstract}


\section{Introduction}

Cement production is vital for global economic development, particularly in supporting infrastructure and construction projects worldwide \cite{wang2023historical, hendriks1998emission}. However, these industrial activities present significant environmental challenges, notably contributing approximately $7\%$ of global carbon emissions \cite{gao2015analysis, korczak2022mitigation}. As nearly 58$\%$ of global cement production occurs in China \cite{asharfi2020quantitative}, and because this region provides the most comprehensive asset-level database of cement facilities \cite{tkachenko2023global}, it offers a statistically representative and data-rich environment for methodological validation. Effective monitoring of these facilities is crucial for environmental management and sustainable development.

Traditional approaches to monitoring cement plants rely on site investigations, which present several limitations in today's rapidly evolving industrial landscape \cite{sawaya2003extending}. These methods prove challenging when monitoring extensive industrial zones, requiring considerable time, resources, and labour whilst failing to provide frequent monitoring capabilities. This has prompted the exploration of remote sensing solutions for more efficient and comprehensive monitoring approaches.

However, detecting cement plants using remote sensing presents several distinct challenges. Firstly, there is a notable scarcity of publicly available datasets for training detection models. Secondly, the intricate and complex background environments of cement plants often contribute to reduced accuracy in detection studies. Furthermore, while traditional optical remote sensing imagery can identify cement plant locations, it proves less effective in determining their operational status.

The clinker generation phase, being the most energy and emission-intensive stage of cement production \cite{worrell2001carbon, ishak2015low}, offers a unique opportunity for detection. During this phase, raw materials undergo heating in kilns at temperatures exceeding 900°C, with limestone ($CaCO_3$) transforming into lime (CaO) and $CO_2$ through calcination at temperatures reaching 1450°C \cite{hendriks2004emission}. These extreme temperatures create distinctive thermal signatures that can be captured through thermal sensors, offering improved insights into operational status assessment \cite{tkachenko2023global, rossi2022detection}.

Recent studies have attempted to leverage thermal signatures for industrial complex detection. For instance, Lie et al. \cite{liu2018identifying} utilised thermal infrared data collected at night to identify industrial heat signatures. Similarly, Ma et al. \cite{ma2018assessing} combined K-Means clustering with heat traces to detect industrial complexes. Building on these approaches, Li et al. \cite{li2024approach} employed Thermal Infrared (TIR) to detect cement plants. However, these methods face challenges in distinguishing industrial complexes from surrounding land cover, due to the complexity of the terrain and the similarity of thermal signatures. 

\begin{figure*}[h!]
    \begin{subfigure}[b]{0.65\linewidth}
        \centering
        \includegraphics[width=\linewidth]{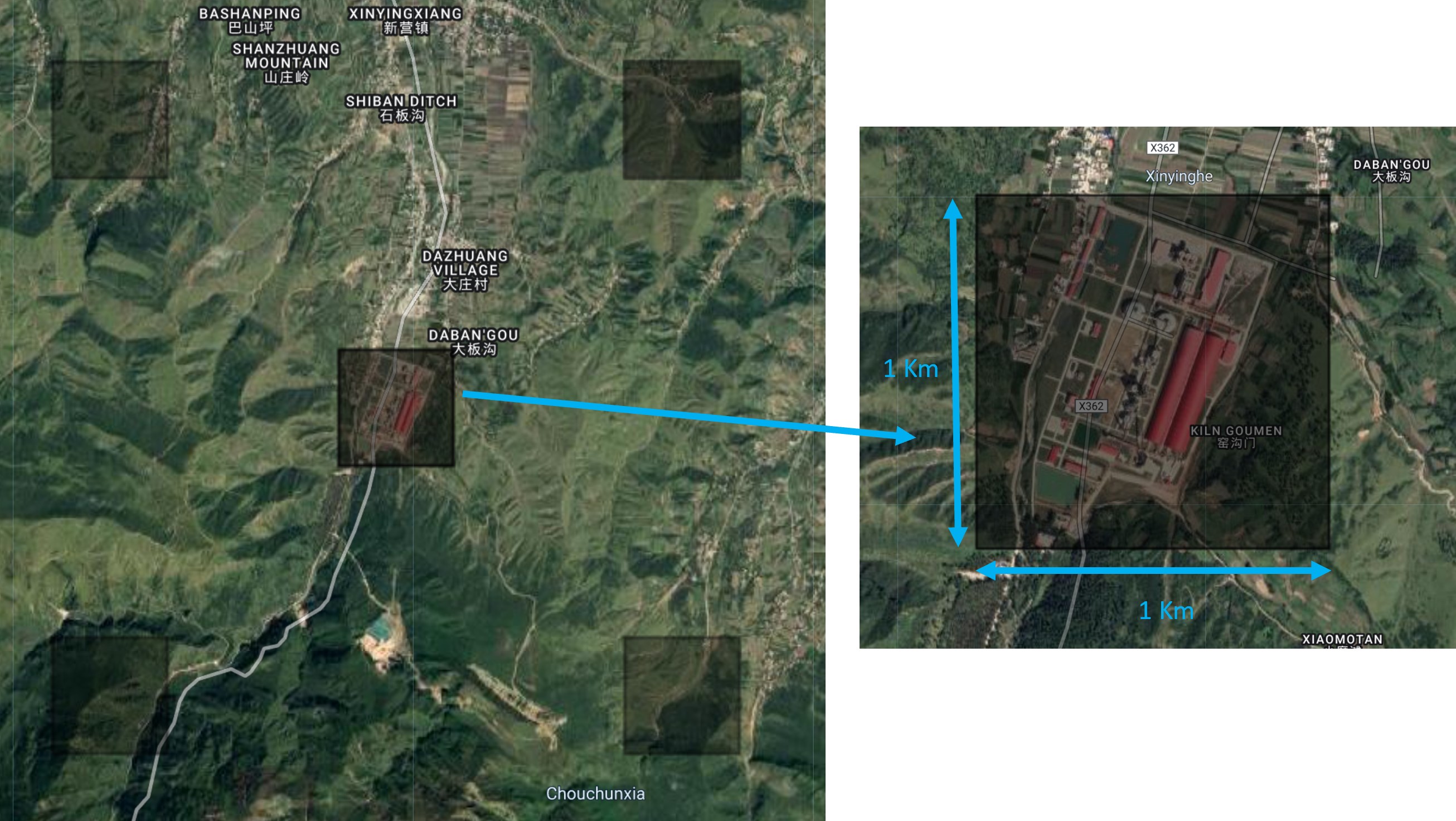}
        \caption{Cement and Surrounding Landcover Chips Sample.}
        \label{fig:cement-chip}
    \end{subfigure}
    \hfill 
    \begin{subfigure}[b]{0.30\linewidth}
        \centering
        \includegraphics[width=\linewidth]{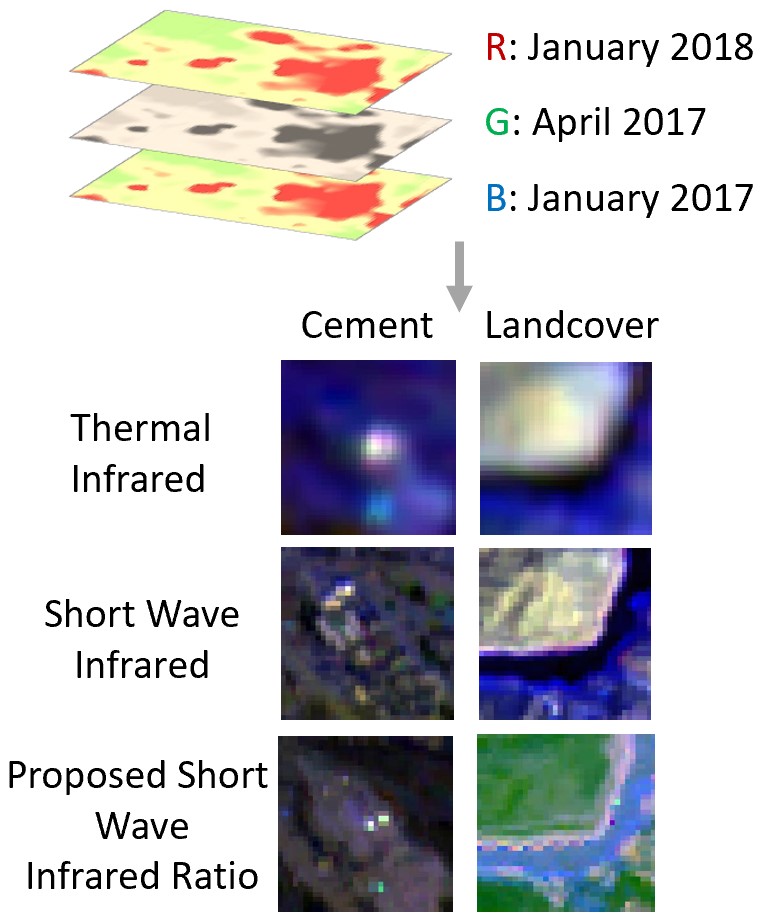}
        \caption{3-Channel 256x256-pixel Images.}
        \label{fig:data-chips}
    \end{subfigure}
    \caption{a) Cement Chip and Surrounding Landcover, b) Each Image is Comprised of 3-Channels, Each Channel Represents a Specific Date: January 2018, April 2017, January 2017.}
    \label{fig:ChinaCementData}
\end{figure*}  

Whilst the existing literature focuses predominantly on the direct physical properties that cause distinct cement heat signatures through the utilisation of TIR, there has been limited investigation into how this excess heat affects the surrounding environment of cement manufacturing facilities. The sustained high temperatures associated with cement production significantly alter soil properties, including moisture content, organic matter composition, and mineralogical characteristics. These alterations manifest in distinctive spectral signatures that can be detected through Short Wave Infrared (SWIR) analysis \cite{ngo2019effective}. Specifically, SWIR bands are particularly sensitive to changes in soil water content, clay mineral composition, and organic matter degradation, all of which are notably affected by prolonged exposure to industrial heat emissions. To bridge this analytical gap, the present work conducts a thorough investigation of these distinct soil signatures. Through the utilisation of SWIR, it is demonstrated that these environmental modifications create more pronounced spectral signatures compared to direct heat signatures, thereby significantly enhancing the model's performance. This novel approach enables more reliable identification of operational cement facilities by examining their long-term environmental impact rather than solely relying on immediate thermal emissions. Accordingly, this study addresses this analytical gap by proposing a physics-informed detection framework that integrates SWIR-derived environmental indicators with TIR thermal cues, enabling more robust and generalisable identification of cement-production facilities.

Although this work focuses on Chinese facilities due to their global dominance and data richness, the proposed framework is sensor-agnostic and readily transferable to other regions and industrial contexts. This study therefore contributes to bridging the methodological gap between thermal- and multi-spectral detection of industrial activity, focusing on China as a globally representative test case.

This paper presents four key contributions: (1) Demonstrate that SWIR bands, particularly the proposed Band 7:6 ratio, provide superior discriminative features for cement plant detection compared to thermal bands alone; (2) Propose FusionNet, an intermediate data fusion architecture that effectively combines thermal (Bands 11, 10) and SWIR (Bands 7, 6, 7:6) features, achieving state-of-the-art performance with $90.6\%$ accuracy; (3) Validate the proposed multi-spectral, physics-informed methodology using a comprehensive dataset of cement plants in China, confirming its effectiveness; (4) Evaluate how embedding differential signal-processing priors, salient features, and wider receptive field affect performance on multispectral and thermal data. Through ablation studies, the contribution of each architectural component is quantified, demonstrating how the resulting backbone enhances the proposed FusionNet relative to traditional models.

\section{Methods}
\subsection{Remotely Sensed Data}

The Global Database of Cement Production Assets (Tkachenko et al. \cite{tkachenko2023global}), a collaboration between the Alan Turing Institute, Green Finance Institute, Satellite Applications Catapult, and the University of Oxford, integrates geospatial data and analytics with financial theory and practice to support sustainable development. This database provides comprehensive asset-level data, including coordinates, ownership information, production type, plant type, production capacity, and the year production commenced. It is a valuable resource for understanding the spatial distribution and characteristics of cement production facilities globally. In this study, the Global Database of Cement Production Assets was utilised to create a shapefile containing detailed information about cement production facilities in China.

\begin{table}[!h]
    \centering
    \caption{Landsat–8 Datasets.}
    \label{tab:Datasets}
    \renewcommand{\arraystretch}{0.6} 
    \scalebox{1}{
    \begin{tabular}{@{\hskip 5pt}c@{\hskip 5pt}c@{\hskip 5pt}c@{\hskip 5pt}c@{\hskip 5pt}c@{\hskip 5pt}} 
        \toprule
        \textbf{Type} & \textbf{Chips Set} & \textbf{Wavelength ($\mu$m)} & \textbf{Cement} & \textbf{Landcover} \\
        \cmidrule{1-5}
        \multirow{2}{*}{Thermal Infrared}
        & Band 10 & 10.60 - 11.19 & 899 & 2,124 \\ 
        & Band 11 & 11.50 - 12.51 & 899 & 3,063 \\
        \cmidrule{1-5}
        \multirow{3}{*}{Shortwave Infrared} 
        & Band 7 & 2.11 - 2.29 & 899 & 2,743 \\
        & Band 6 & 1.57 - 1.65 & 899 & 3,069 \\
        & Band 7:6 & N/A & 899 & 2,807 \\
        \bottomrule
        \vspace{-0.3cm}
    \end{tabular}
    }
\end{table}

\begin{figure*}[t!]
    \centering
    \begin{subfigure}[b]{0.55\linewidth}
        \centering
        \includegraphics[width=\linewidth]{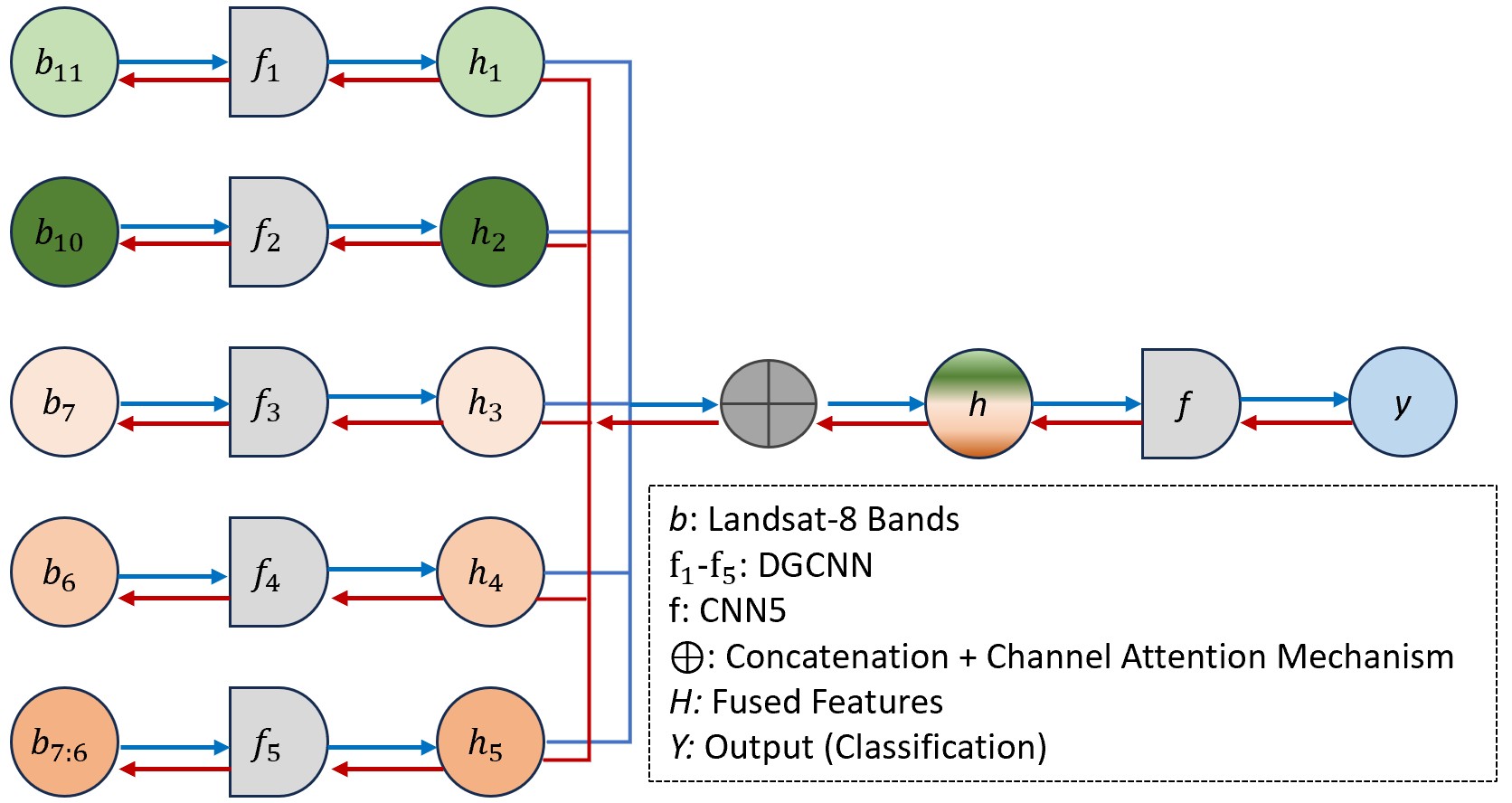}
        \caption{FusionNet.}
        \label{fig:fusionnet}
        \vspace{0.5cm} 
    \end{subfigure}
    \hfill 
    \begin{subfigure}[b]{0.44\linewidth}
        \centering
        \includegraphics[width=\linewidth]{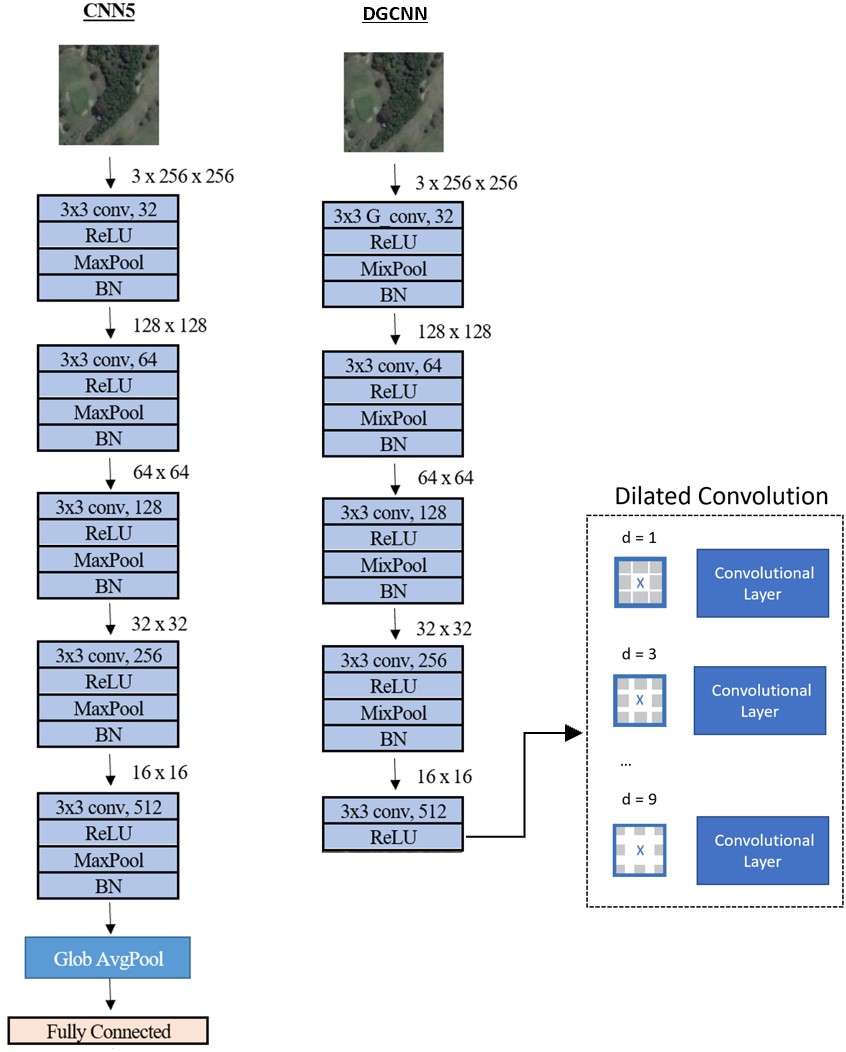}
        \caption{Backbone Networks.}
        \label{fig:backbones}
        \vspace{0.5cm} 
    \end{subfigure}
    \begin{subfigure}[b]{0.70\linewidth}
        \centering
        \includegraphics[width=\linewidth]{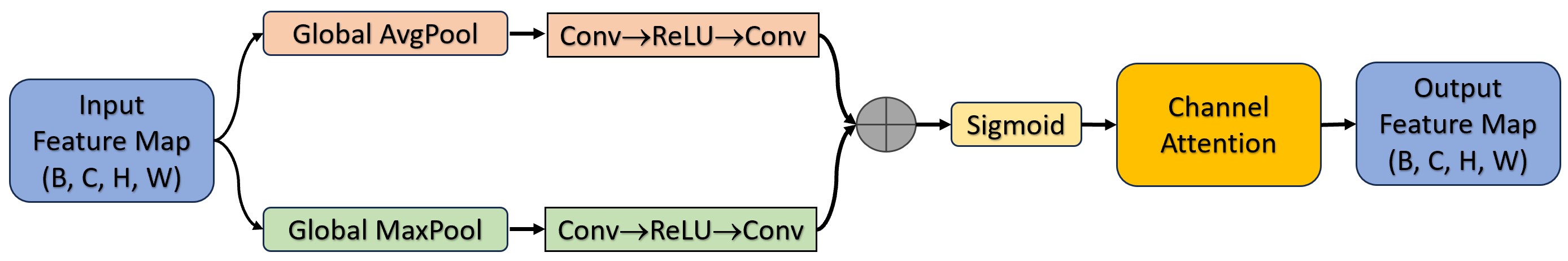}
        \caption{Channel Attention Mechanism.}
        \label{fig:channelattention}
        \vspace{0.3cm} 
    \end{subfigure}
    \caption{a) FusionNet: TIR (Bands 11-10) and SWIR (Bands 7-6 and Ratio 7:6), b) Backbone Networks: $f_1-f_5$: 5 Convolutional Layer Network (DGCNN) with an Initial Parameterised Gabor Convolutional Layer, a Mixture of Average and Maximum Pooling, and a Dilated Convolutional Layer; f: 5 Convolutional Layer Network (CNN5), c) Channel Attention Mechanism.}
    \vspace{-0.5cm}
    \label{fig:combined-figures}
\end{figure*}

Utilising the shapefile, 1x1 $km^2$ grid regions classified as Cement Plant and Landcover chips were extracted from Google Earth Engine using Landsat 8 data (Figure \ref{fig:cement-chip}). The analysis focused on several key datasets derived from Landsat 8. TIR bands (Bands 10 and 11) were utilised to detect heat emissions characteristic of cement plants. SWIR bands (Bands 6 and 7) were utilised to identify materials and moisture content, which are crucial for distinguishing cement plants from other land cover types. Combining Bands 6 and 7 further enhanced the detection and classification accuracy (Figure \ref{fig:data-chips}). Table \ref{tab:Datasets} shows the proposed five datasets.

\textbf{Thermal Infrared Signature}: The potential of using Landsat 8 thermal Bands 10 and 11 for identifying cement assets was assessed by applying corrections for top-of-atmosphere and brightness temperature. Seasonal analysis revealed that winter months, particularly January, provided the most effective thermal differentiation between cement plants and their surroundings. April also proved useful, as the surrounding land began to warm up while the thermal signatures of the plants remained consistent. This seasonal variability aids the model in distinguishing the continuous thermal signature of cement plants from transient hotspots. Consequently, all images used in the analysis are 3-channel images, with each channel representing January 2017, April 2017, and January 2018.

\textbf{Short Wave Infrared Signature:} SWIR wavelengths, typically ranging from 0.9 to 2.5 microns, are effective for distinguishing cement plants from surrounding land due to their sensitivity to mineral composition and moisture content. This study examines how the Band 7:6 ratio from Landsat 8 can enhance the contrast between cement plants and other land cover types. The effectiveness of this approach can be attributed to several key characteristics:

\begin{enumerate}
    \item \textit{Spectral Reflectance Properties:} Cement plants show distinct reflectance characteristics in SWIR Bands 7 (2.11–2.29 µm) and 6 (1.57–1.65 µm). The Band 7:6 ratio highlights these contrasts, making them stand out against vegetation, soil, and other natural features.
    \item \textit{Moisture Content Sensitivity and Enhanced Detection:} Cement plants, with lower moisture content than vegetation and soil, are effectively highlighted by the Band 7:6 ratio. This ratio is sensitive to moisture levels and accentuates the unique spectral signature of cement plants, facilitating their classification and segregation from other land cover types.
    \item \textit{Reduction of Atmospheric Effects:} The Band 7:6 ratio reduces atmospheric water vapour effects by normalising similar perturbations in this spectral region, enhancing the reliability of cement plant detection by minimising atmospheric interference.
\end{enumerate}

Cement plants exhibit distinctive Band~7:6 ratios that differ from surrounding features, owing to their unique composition and moisture characteristics. These spectral differences enable reliable segregation from natural features such as vegetation. As a compact descriptive check, the absolute differences between cement and landcover class means of per-chip mean pixel values were 18.74 for Band~6, 19.61 for Band~7, and 23.75 for Band~7:6, indicating that the proposed ratio provides the largest separation among the SWIR representations.

\subsection{Models}

\noindent\textbf{FusionNet.} This work proposes an intermediate data fusion model, termed FusionNet, to explore its ability to preserve and combine distinct spectral features at a more abstract level. The model is trained on Landsat 8 data, including Bands 11, 10, 7, 6, and the Band ratio 7:6. The architecture comprises two main components. Firstly, a unimodal feature extraction stage utilises five convolutional neural networks ($f_1, f_2, …, f_5$), each trained with one of the input bands or the derived band ratio. Each network extracts unimodal features ($h_1, h_2, …, h_5$) using a specialised convolutional backbone (DGCNN). These backbones incorporate an initial differential Gabor-parameterised convolutional layer, a mixture of maximum and average pooling, and an average dilated convolutional layer, enabling the capture of fine-grained and global spatial features (Voulgaris et al. \cite{voulgaris2023seasonal}).

Secondly, the fusion module combines the unimodal features through concatenation, followed by a channel attention mechanism that identifies the most informative features during training and enhances their contribution. The resulting fused features (h) are then passed to a five-layer convolutional neural network (CNN5), which processes the data to make final predictions, referred to as the target outputs (y).

The proposed FusionNet framework leverages the complementary strengths of unimodal feature extraction and multi-spectral fusion. Figure \ref{fig:combined-figures} provides an overview of the model, including the DGCNN and CNN5 backbones, illustrating the flow from Landsat 8 spectral inputs to the final predictions.

\subsubsection{\textbf{CNN5}}

The first model under consideration is comprised of five hidden convolutional layers. Each convolutional layer includes a convolutional layer, followed by an activation function (ReLU), a maximum pooling layer and a batch normalisation. Furthermore, the last convolutional layer is connected to a global average pooling layer. Global pooling was first mentioned in the work of Lin et al. \cite{lin2013network} and became mainstream with ResNet by He et al. \cite{he2016deep}.  

\subsubsection{\textbf{Dilated Gabor Mix Pool CNN (DGCNN)}}

Dilated convolution was first introduced by Chen et al. \cite{chen2014semantic, chen2017deeplab} as a means of increasing the receptive field for semantic segmentation tasks. According to Wei et al. \cite{wei2018revisiting}, convolutional kernel receptive fields are enlarged when employing varying dilation rates, enabling the transfer of surrounding discriminative information to key scene regions.

In prior work \cite{voulgaris2023seasonal}, we introduced a model comprising a convolutional layer parameterised by trainable Gabor functions, a mixture of maximum and average pooling (MixPool), and an averaged dilated convolutional layer. This design embedded differential signal processing priors, enabled salient feature extraction with a wider receptive field, and demonstrated improved performance under seasonal domain shifts.

The present study investigated how integrating these blocks into the proposed intermediate data fusion framework, FusionNet, affects classification performance when applied to LandSat 8 multi-spectral data. Specifically, the model was evaluated across TIR (Bands 10 and 11), SWIR (Bands 6 and 7), a geological SWIR ratio (Band 7:6), and a fused dataset comprising all five inputs. The objective was to assess whether embedding differential signal-driven priors with wider receptive fields yielded measurable advantages over conventional CNN architectures in multi-spectral fusion scenarios.

Specifically, for the differential Gabor-parameterised convolutional layer, a bank of Gabor filters was used: 
\begin{align}
        g(x, y, \omega, \theta, \psi, \sigma) = \text{exp}(- \frac{x'^2 + y'^2}{2\sigma^2}) \text{cos}(\omega x' + \psi)\\
                x' = x\text{cos}\theta + y\text{sin}\theta;
        \quad
        y' = -x\text{sin}\theta + y\text{cos}\theta,\nonumber
\end{align}

\noindent where $(x, y)$ stands for the pixel spatial domain position, $\theta$ the filter orientation, $\sigma$ the standard deviation, and $\omega$ the frequency. A bank of Gabor filters with frequencies $\omega_n$ and orientations $\theta_m$ was used:  
\begin{equation}
    \omega_n = \frac{\pi}{2}\sqrt{2}^{-(n-1)} \quad{\text{where}, n = 1, 2, ..., 5}
\end{equation}
\begin{equation}
    \theta_m = \frac{\pi}{8}(m-1) \quad{\text{where}, m = 1, 2, ..., 8}.
\end{equation}

Gabor layer weights were initialised by setting the standard deviation to $\sigma \approx \frac{\pi}{\omega}$. Furthermore, $\psi$ was set by uniform distribution $\text{Unif.}(0, \pi)$. Note that the Gabor function parameters were updated during backpropagation. 

In addition, due to the information complexity in satellite images, a method that combines maximum and average pooling was applied. Hence on the lowest layer, Maximum pooling was applied, and as incremented to higher layers, it was decreased by 0.2: 
\begin{equation}
    \label{eq:MixPool}
    f_{\text{mix}}(x) = \alpha_l \cdot f_{\text{max}}(x) + (1 - \alpha_l) \cdot f_{\text{avg}}(x),
\end{equation}
\noindent where scalar mixing portion $\alpha_l\in[0,1]$ indicates the max and average combination per layer \textit{l}.

As an evolution of our previous architecture, similar to \cite{wei2018revisiting}, a dilated convolutional layer was added between the last convolutional layer and the global average pooling. Specifically, convolutional blocks with multiple dilated rates (i.e. d = 1, 3, 6, 9) were appended to the final convolutional layer, thus localising scene-related regions observed by different receptive fields. Using high dilation rates (i.e. d = 9) can cause inaccuracies by mistakenly highlighting scene-irrelevant regions. To avoid such scenarios, equation \ref{eq:Dil} was used, where the average over the localisation maps $H_i$ (i.e. i = 3, 6, 9) generated by different dilated convolutional blocks was summed to the localisation map $H_0$ of the convolutional block with dilation d = 1.
\vspace{-5pt}
\begin{equation}
    \label{eq:Dil}
    H = H_o + \frac{1}{n_d} \sum_{n=1}^{n_d}H_i
\end{equation}

\subsubsection{\textbf{Channel Attention Mechanism}}

\noindent To enhance the discriminative capacity of the fused feature representation, a lightweight channel attention module inspired by global context modelling is integrated. Given the concatenated multi-branch output $\mathbf{F} \in \mathbb{R}^{B \times C \times H \times W}$, where $C = 5 \times \text{CNN}_{\text{out}}$, the attention mechanism adaptively recalibrates channel-wise responses by exploiting both average and maximum pooled statistics.

\begin{algorithm}[h]
\caption{FusionNet — Initialisation and Forward Pass}
\label{alg:fusionnet}
\begin{algorithmic}[1]

\Require Batch inputs 
\Statex \quad $X_{b10},X_{b11},X_{b7},X_{b6},X_{b7:6}\in\mathbb{R}^{B\times C\times H\times W}$
\Ensure Logits $Z\in\mathbb{R}^{B\times 2}$, class probabilities $\hat{P}\in\mathbb{R}^{B\times 2}$ and predicted labels 
\Statex \quad $Y^\ast\in\{0,1\}^B$

\Statex
\State \textbf{Initialisation}
\State Create five independent DGCNN branches (no weight sharing):
\State \quad $f_{b10}\gets\textsc{DGCNN}()$ \Comment{Dilated Gabor Parameterised MixPool CNN: GaborConv $\rightarrow$ ReLU $\rightarrow$ MixPool $\rightarrow$ BN $\rightarrow$ Conv blocks $\rightarrow$ Dilated Conv Block}
\State \quad $f_{b11}\gets\textsc{DGCNN}()$
\State \quad $f_{b7} \gets\textsc{DGCNN}()$
\State \quad $f_{b6} \gets\textsc{DGCNN}()$
\State \quad $f_{b7:6}\gets\textsc{DGCNN}()$ \Comment{branch for Band 7:6 ratio}
\State Define channel attention module: $\textsc{CA}\gets\textsc{ChannelAttention}()$ \Comment{CA: GlobalAvg/MaxPool $\rightarrow$ shared MLP (Conv$\to$ReLU$\to$Conv) $\rightarrow$ sigmoid}
\State Define fusion classifier: $f_{\text{fusion}}\gets\textsc{CNN5}()$ \Comment{CNN5 + GlobalAvgPool + FC}

\Statex
\Function{ForwardBatch}{$X_{b10},X_{b11},X_{b7},X_{b6},X_{b7:6}$}
    \Comment{Unimodal feature extraction; each branch returns $h_{*}\in\mathbb{R}^{B\times C_i\times h\times w}$}
    \State $h_{b10}\gets f_{b10}(X_{b10})$
    \State $h_{b11}\gets f_{b11}(X_{b11})$
    \State $h_{b7} \gets f_{b7}(X_{b7})$
    \State $h_{b6} \gets f_{b6}(X_{b6})$
    \State $h_{b7:6}\gets f_{b7:6}(X_{b7:6})$

    \Statex
    \Comment{Assume all $h_{*}$ have identical spatial dims $(h,w)$; otherwise up/downsample to match}
    \State $H\gets\textsf{concat}(h_{b10},h_{b11},h_{b7},h_{b6},h_{b7:6})$ \Comment{$H\in\mathbb{R}^{B\times C_{\text{concat}}\times h\times w}$}

    \State $u_{\text{avg}}\gets\text{GlobalAvgPool}(H)$ \Comment{$\in\mathbb{R}^{B\times C_{\text{concat}}\times1\times1}$}
    \State $u_{\text{max}}\gets\text{GlobalMaxPool}(H)$
    \State $m_{\text{avg}}\gets\text{MLP}(u_{\text{avg}})$ \Comment{MLP = Conv $\rightarrow$ ReLU $\rightarrow$ Conv; preserves channel dim}
    \State $m_{\text{max}}\gets\text{MLP}(u_{\text{max}})$
    \State $M\gets\sigma(m_{\text{avg}}+m_{\text{max}})$ \Comment{$M\in[0,1]^{B\times C_{\text{concat}}\times1\times1}$}

    \State $H'\gets H\odot M + H$ \Comment{channel-wise recalibration with residual}

    \State $Z\gets f_{\text{fusion}}(H')$ \Comment{logits produced by CNN5 (global pooling + FC)}
    \State $\hat{P}\gets\text{softmax}(Z)$
    \State $Y^\ast\gets\arg\max(\hat{P},\text{axis}=1)$

    \State \Return $Z,\,\hat{P},\,Y^\ast$
\EndFunction

\end{algorithmic}
\end{algorithm}

\setlength{\textfloatsep}{8pt}
Adaptive average pooling and adaptive maximum pooling are independently applied to $\mathbf{F}$, producing two descriptors of shape $\mathbb{R}^{B \times C \times 1 \times 1}$. These descriptors are passed through a shared multi-layer perceptron (MLP) comprising two $1 \times 1$ convolutional layers with a ReLU activation in between. The outputs are summed and activated via a sigmoid function to yield channel-wise attention weights $\mathbf{A} \in \mathbb{R}^{B \times C \times 1 \times 1}$.

The refined feature map is computed as:
\[
 F_{\text{refined}} = F + F \odot A
\]
where $\odot$ denotes element-wise multiplication, and the residual connection preserves robustness under noisy or adversarial conditions.

This mechanism enables the network to selectively emphasise informative channels across spectral branches, suppressing redundancy and enhancing inter-branch synergy. Unlike spatial attention, which operates locally, the channel attention module captures global semantic dependencies, with minimal computational overhead, making it well-suited for high-dimensional fusion scenarios.

Algorithm \ref{alg:fusionnet} summarises the complete FusionNet pipeline, detailing branch initialisation, unimodal feature extraction, channel-wise attention recalibration, and final classification.

\subsubsection{\textbf{Transformer-Based Baselines}}

To evaluate performance against attention based architectures, two lightweight Vision Transformer \cite{dosovitskiy2020image} variants were implemented: ViT\_Tiny and ViT\_Compact. Both models employ linear patch embeddings, learnable positional encodings, and multi-head self-attention blocks. Each model uses 4 Transformer layers with a hidden embedding dimension of 128, 4 attention heads, and a patch size of 16; ViT\_Compact adopts a reduced MLP expansion compared to ViT\_Tiny. No external pretraining was used to ensure a fair comparison under identical data conditions.

\begin{table}[h]
    \centering
    \caption{Training Configuration.}
    \label{tab:training_config}
    \renewcommand{\arraystretch}{1.2}
    \setlength{\tabcolsep}{9pt}
    \scalebox{1.15}{
    \begin{tabular}{ll}
        \toprule
        Parameter & Value \\
        \midrule
        Epochs & 150 \\
        Batch size & 90 \\
        Optimiser & Adam \\
        Learning rate & $1 \times 10^{-4}$ \\
        Loss & Cross-Entropy \\
        Class weights & [3, 1] \\
        Data augmentation & Rotation (0.9), H/V flip (0.5 / 0.1) \\
        Repeats & 5 independent splits \\
        \bottomrule
        \vspace{-0.8cm}
    \end{tabular}
    }
\end{table}

\subsection{Model Training}

All models were trained for 150 epochs (based on when train/validation results stopped improving), using Adam Optimiser and a weighted Cross Entropy loss, defined as $\mathcal{L} = - \sum_{c=1}^{C} w_c\, y_c \log(\hat{y}_c)$, where $w_c$ is the class weight and $\hat{y}_c$ the predicted probability for class $c$, as summarised in Table~\ref{tab:training_config}. The dataset split was 80\% for train/validation purposes (out of which 80\% train and 20\% validation) and 20\% for testing, with exact sample counts reported in Table~\ref{tab:dataset_split}. To tackle class imbalance and ensure classifiers were not biased towards the majority class, stratification was applied to the data. Moreover, since landcover is overrepresented compared to cement class in a ratio 4:1, class weighting [3, 1] was used. The only data augmentations applied during training were geometric transformations, namely, rotation with 90\% probability, horizontal and vertical flip with 50\% and 10\% probabilities respectively. All models were trained over five independent data splits.

\begin{table}[h]
    \centering
    \caption{Dataset Composition and Train/Validation/Test Split per Spectral Configuration.}
    \label{tab:dataset_split}
    \renewcommand{\arraystretch}{1.6}
    \setlength{\tabcolsep}{10pt}
    \scalebox{1.15}{
    \begin{tabular}{lrccc}
        \toprule
        Dataset & Total & Train & Val & Test \\
        \midrule
        Band 10 (TIR)  & 3,023 & 1,935 & 484 & 605 \\
        Band 11 (TIR)  & 3,962 & 2,536 & 634 & 792 \\
        Band 7 (SWIR)  & 3,642 & 2,331 & 583 & 728 \\
        Band 6 (SWIR)  & 3,968 & 2,540 & 635 & 794 \\
        Band 7:6 (SWIR) & 3,706 & 2,372 & 593 & 741 \\
        \bottomrule
        \vspace{-0.8cm}
    \end{tabular}
    }
\end{table}

\subsection{Performance Metrics}

To quantify model performance, four standard classification metrics were used: accuracy, precision, recall, and F1‑score. Accuracy, defined as the proportion of correctly classified samples, provides a general indication of performance but can be misleading when class distributions are imbalanced. For this reason, precision and recall were additionally reported. Precision measures the proportion of predicted positives that are correct, while recall captures the proportion of actual positives that are successfully identified. As these metrics often trade off against one another, the F1‑score, their harmonic mean, offers a balanced measure of class‑specific performance.

All metrics were computed from the confusion matrix using the counts of true positives (TP), true negatives (TN), false positives (FP), and false negatives (FN):

\begin{equation}
\begin{aligned}
\text{Accuracy} &= \frac{TP + TN}{TP + TN + FP + FN}, \\
\text{Precision} &= \frac{TP}{TP + FP}, \\
\text{Recall} &= \frac{TP}{TP + FN}, \\
F_{1} &= \frac{2 \cdot \text{Precision} \cdot \text{Recall}}
{\text{Precision} + \text{Recall}}.
\end{aligned}
\end{equation}

\section{Experimental Results}

\subsection{Ablation Studies}

A series of ablation studies was conducted to evaluate how embedding differential signal processing priors, a mixed pooling strategy that enables the retention of both strong activations and contextual information, and an expanded receptive field affect model performance, both individually and combined. Notably, as standard deviations were consistently $\leq$0.01$\%$ across all experiments, only mean accuracy is reported for clarity.

\noindent\textbf{GCNN5}. The impact of embedding differential signal processing priors via learnable Gabor filters, optimised through backpropagation is examined. Parameterised Gabor convolutional layers enable the capture of orientation and frequency-specific features, providing robust low-level representations. This is particularly beneficial in remote sensing tasks, where texture and edge patterns are critical for accurate classification.

\noindent\textbf{MPCNN5}. The effect of combining maximum pooling, which emphasises strong activations, with average pooling, which preserves contextual information is assessed. This mixed pooling strategy reduces overfitting and improves generalisation by balancing local saliency with broader spatial context.

\noindent\textbf{DCNN5}. The role of averaged dilated convolutions in expanding the receptive field without increasing the number of parameters is evaluated. This facilitates the capture of multi-scale spatial dependencies, which are essential for distinguishing complex structures such as cement plants, and contributes to improved classification performance.

\begin{table}[!h]
    \centering
    \caption{Ablation Study of the Proposed Layers. Results are Reported as Mean Accuracy ($\%$) Over Five Independent Splits.}
    \label{tab:AvgAcc}
    \renewcommand{\arraystretch}{1.2} 
    \scalebox{1.15}{
    \begin{tabular}{@{\hskip 6pt}l@{\hskip 6pt}c@{\hskip 6pt}c@{\hskip 6pt}c@{\hskip 6pt}c@{\hskip 6pt}c@{\hskip 6pt}} 
        \toprule
        \textbf{Models} & \textbf{Band 11} & \textbf{Band 10} & \textbf{Band 7} & \textbf{Band 6} & \textbf{Band 7:6} \\
        \cmidrule{1-6}
        CNN5 & 79.5 & 76.9 & 82.6 & 82.3 & 83.3 \\
        MPCNN5 & 80.7 & 77.6 & 82.8 & 83.2 & 84.4 \\  
        GCNN5 & 82.5 & 80.7 & 85.8 & 84.4 & 85.4 \\
        DCNN5 & 83.6 & 81.9 & 86.4 & 85.6 & 86.9 \\
        DGCNN & \textbf{84.2} & \textbf{83.6} & \textbf{87.7} & \textbf{86.2} & \textbf{88.7} \\
        \bottomrule
        \vspace{-0.3cm}
    \end{tabular}
    }
\end{table}

Table~\ref{tab:AvgAcc} presents the results of an ablation study designed to isolate the contribution of each proposed layers within the DGCNN backbone. The baseline CNN5 yields the lowest performance across all spectral inputs, with accuracy ranging from 76.9$\%$ (Band 10) to 83.3$\%$ (Band 7:6), reflecting limited capacity for capturing spatial and spectral complexity.

Introducing mixed pooling (MPCNN5) offers modest gains, particularly in SWIR bands, suggesting improved generalisation through the retention of both strong activations and contextual information. Embedding Gabor-parameterised convolutions (GCNN5) leads to more substantial improvements, especially in Band 7 and Band 11, where orientation- and frequency-sensitive features enhance texture discrimination. Averaged dilated convolutions (DCNN5) further elevate performance by expanding the receptive field, enabling the model to capture multi-scale spatial dependencies critical for distinguishing cement plant structures.

\addtocounter{figure}{1}
\begin{figure*}[!b]
    \centering
    \begin{subfigure}[b]{0.45\linewidth}
        \centering
        \includegraphics[width=\linewidth]{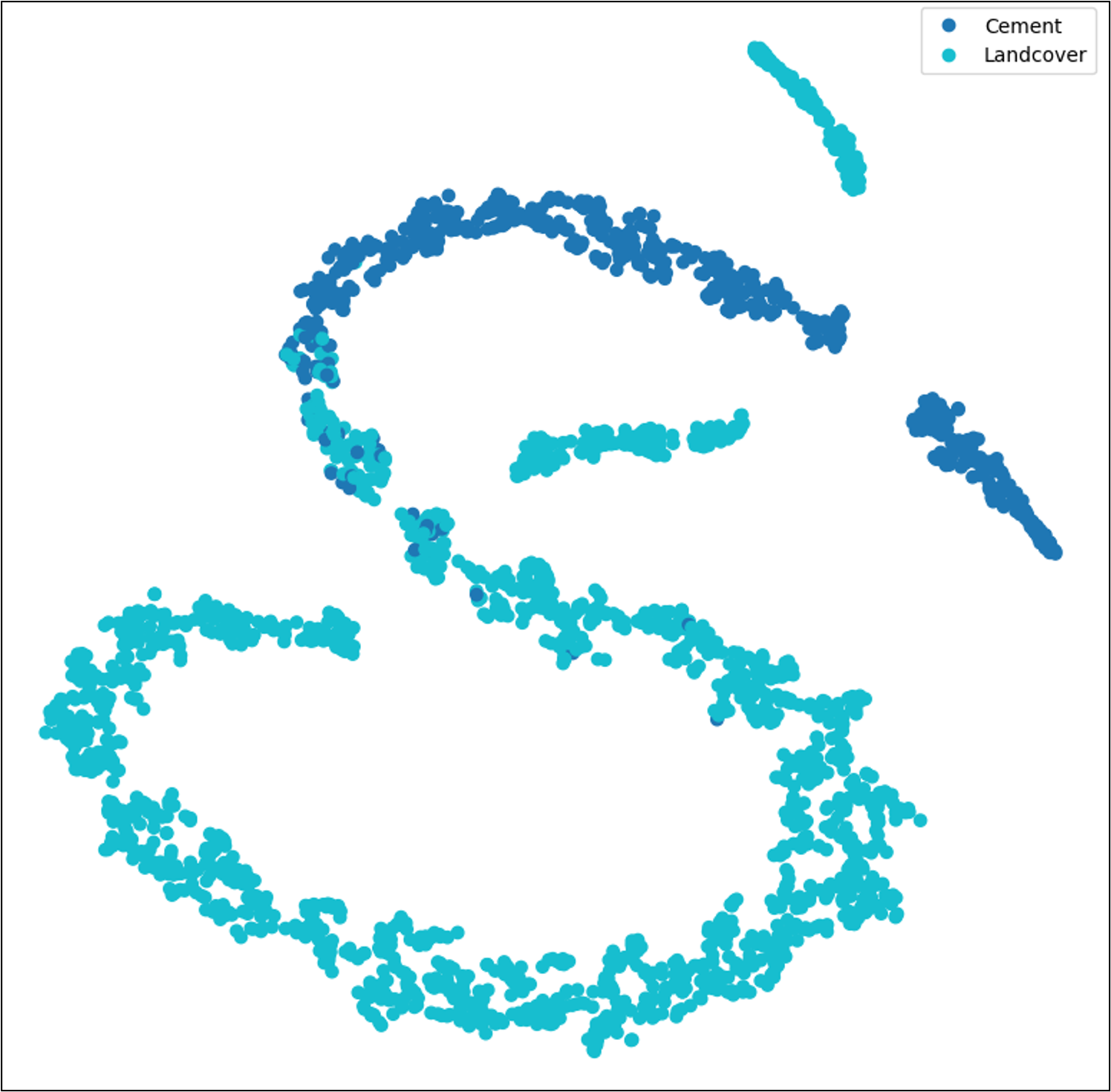}
        \caption{CNN5 t-SNE.}
        \label{fig:cnn5tsne}
    \end{subfigure}
    \hfill 
    \begin{subfigure}[b]{0.45\linewidth}
        \centering
        \includegraphics[width=\linewidth]{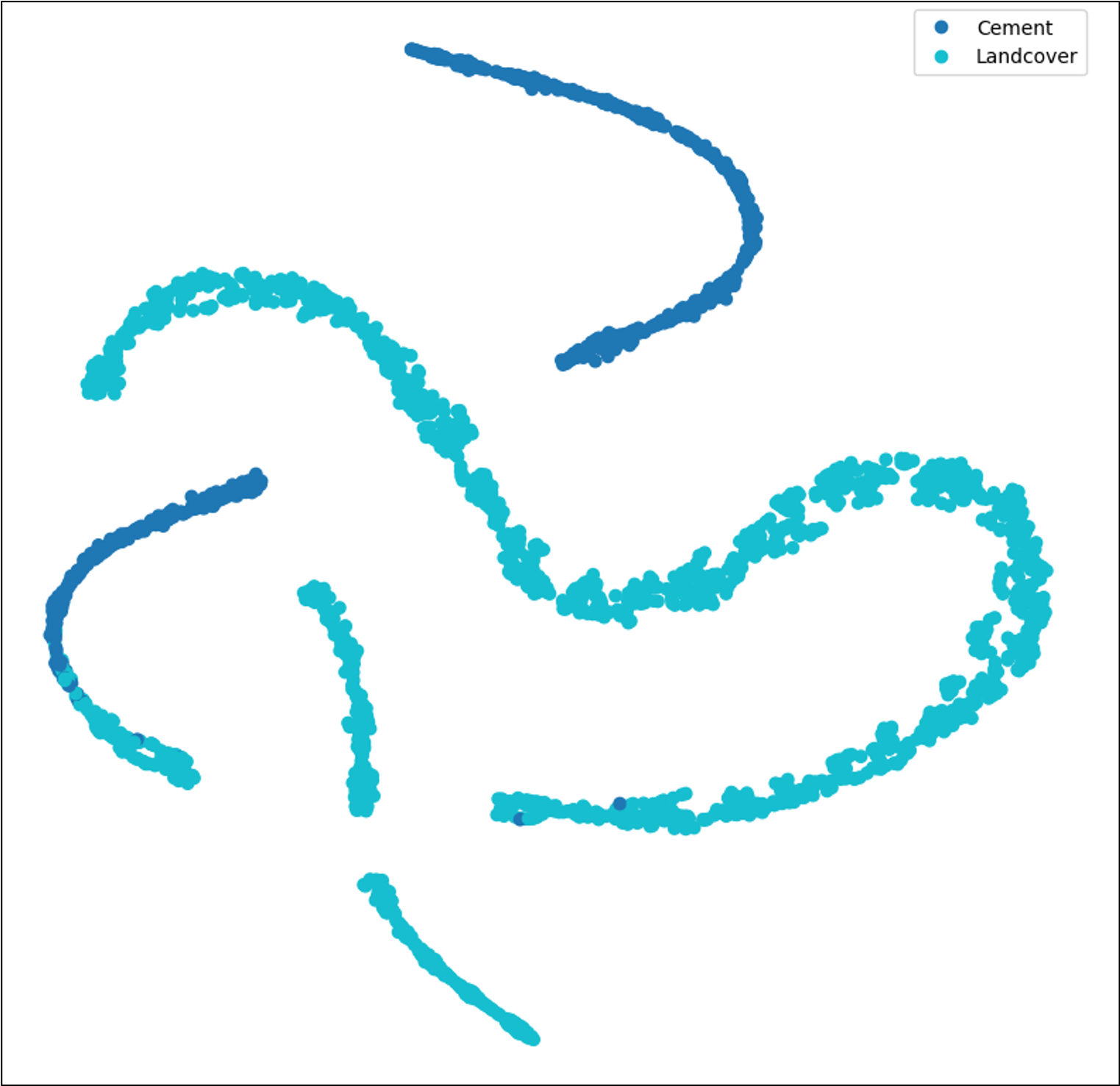}
        \caption{DGCNN t-SNE.}
        \label{fig:dgcnn5tsne}
    \end{subfigure}
    \caption{t-Distributed Stochastic Neighbour Embedding (t-SNE) a) CNN5, b) DGCNN per Cement and Landcover Classes. The Proposed DGCNN Model, Yields a Clearer Separation Between Classes Compared to Traditional CNNs.}
    \label{fig:tsne}
\end{figure*}
\addtocounter{figure}{-2}

The full integration of all three components in DGCNN yields the highest accuracy across all bands, with peak performance on the geological SWIR ratio (Band 7:6) at 88.7$\%$. These results confirm that each proposed layer contributes positively to classification performance, and their combination produces a synergistic effect. Notably, the consistent gains across both thermal and SWIR domains underscore the robustness of the proposed backbone in handling spectral variability and spatial complexity inherent to remote sensing tasks.

\begin{table}[!h]
    \centering
    \caption{Ablation Study of the Proposed Backbone vs a Conventional Convolutional Backbone on the Proposed FusionNet. Results are Reported as Mean Accuracy ($\%$) Over Five Independent Splits.}
    \label{tab:FusionNetAvgAcc}
    \renewcommand{\arraystretch}{1.0} 
    \scalebox{1.085}{
    \begin{tabular}{@{\hskip 5pt}l@{\hskip 5pt}c@{\hskip 5pt}c@{\hskip 5pt}c@{\hskip 5pt}c@{\hskip 5pt}c@{\hskip 5pt}} 
        \toprule
        \textbf{Models} & \textbf{Band 11} & \textbf{Band 10} & \textbf{Band 7} & \textbf{Band 6} & \textbf{Band 7:6} \\
        \cmidrule{1-6}
        \footnotesize FusionNet$^{\text{CNN5}}$ & 83.1 & 82.9 & 87.4 & 86.8 & 87.8 \\ 
        \footnotesize FusionNet$^{\text{DGCNN}}$ & \textbf{86.9} & \textbf{85.7} & \textbf{89.6} & \textbf{88.4} & \textbf{90.6} \\
        \bottomrule
        \vspace{-0.3cm}
    \end{tabular}
    }
\end{table}

Table~\ref{tab:FusionNetAvgAcc} presents the ablation study results, comparing the performance of FusionNet when equipped with a conventional convolutional backbone (CNN5) versus the proposed DGCNN. The models were evaluated across five spectral inputs derived from Landsat 8: TIR (Bands 11 and 10), SWIR (Bands 7 and 6), and the geological SWIR ratio (Band 7:6).

Across all spectral bands, FusionNet with the proposed DGCNN backbone, demonstrated accuracy improvements ranging from 1.6$\%$ to 3.8$\%$. The most pronounced gains are observed in Band 11 and the SWIR ratio (Band 7:6), where the proposed backbone achieves 86.9$\%$ and 90.6$\%$ accuracy respectively, compared to 83.1$\%$ and 87.8$\%$ with CNN5. These results underscore the effectiveness of embedding differential Gabor-parameterised convolutions, mixed pooling, and averaged dilated receptive fields in enhancing feature extraction.

Notably, the improvements are consistent across both thermal and SWIR domains, suggesting that the proposed backbone is not only more expressive but also more resilient to spectral variability. The superior performance on the SWIR ratio further validates the utility of spectral combinations sensitive to surface alterations, which complement the direct thermal cues captured by TIR bands. Overall, the ablation confirms that proposed architecture yield measurable gains in cement plant detection accuracy across diverse spectral inputs.

\subsection{FusionNet Backbone Ablation (DGCNN vs CNN5)}

A series of experiments is conducted to evaluate how embedding differential signal processing priors, salient feature extraction, and a wider receptive field compares with traditional convolutional architectures. In addition, the impact of direct thermal properties versus indirect indicators, such as soil composition alteration, is assessed in the context of cement plant detection relative to the surrounding land cover.

\begin{figure}[!h]
    \centering
    \includegraphics[width=1\linewidth]{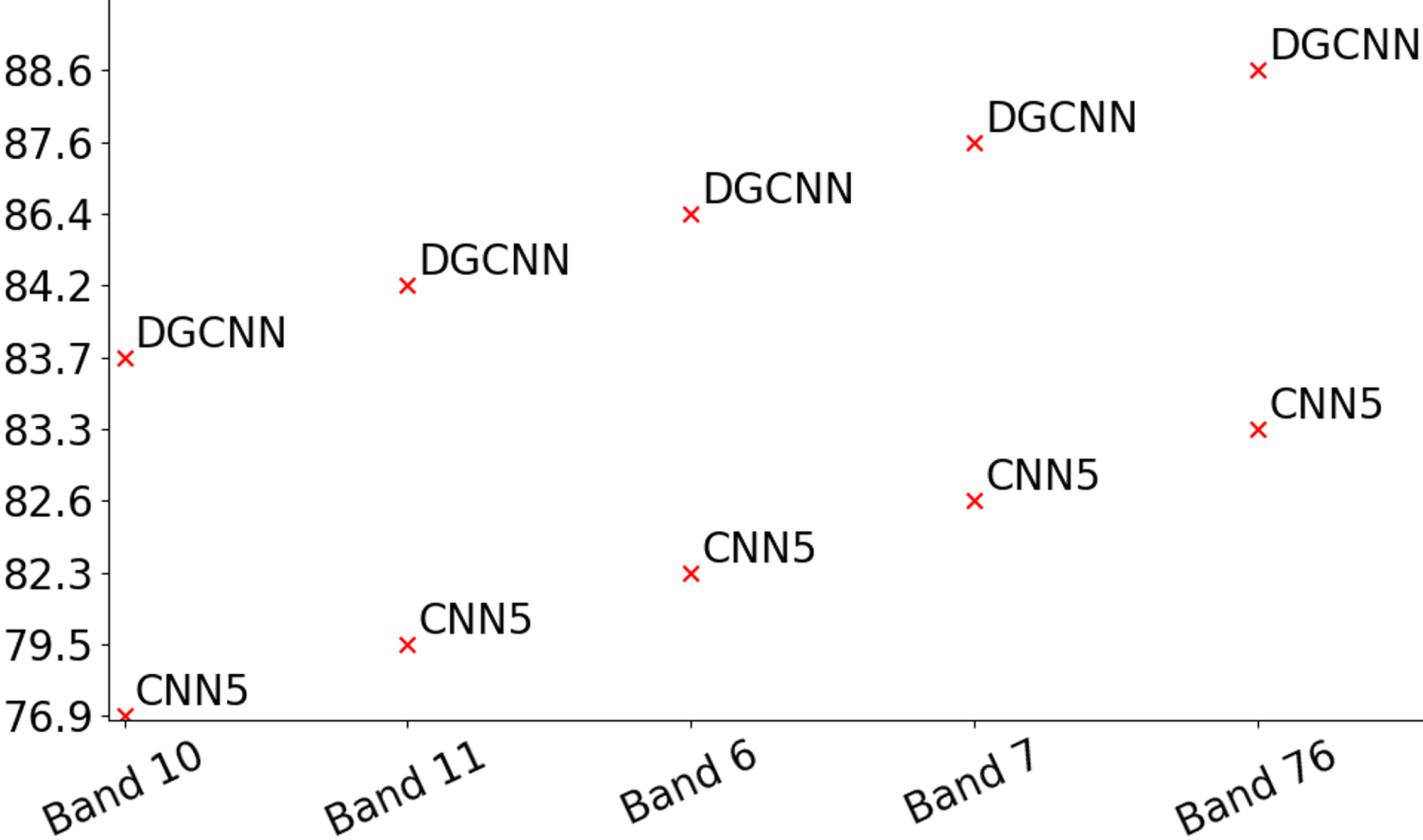}
    \caption{CNN5 vs DGCNN Average Accuracy ($\%$) Scores Across TIR (Bands 11–10) and SWIR (Bands 7–6 and Ratio 7:6) Inputs, Evaluated per Cement and Landcover Classes. The Proposed DGCNN Outperforms the Conventional Convolutional Backbone in Both Thermal and SWIR Datasets, with the Proposed Ratio Yielding the Highest Overall Accuracy.}
    \vspace{-0.2cm}
    \label{fig:CNN4DGCNN5}
\end{figure}
\addtocounter{figure}{1}

Figure~\ref{fig:CNN4DGCNN5} presents the average classification accuracy of the baseline CNN5 and the proposed DGCNN across Landsat 8 spectral inputs: TIR (Bands 10 and 11), SWIR (Bands 6 and 7), and the geological SWIR ratio (Band 7:6). Three key observations emerge. First, DGCNN consistently outperforms CNN5 across all spectral inputs, underscoring the benefits of embedding differential signal-driven priors, salient feature extraction, and expanding the receptive field. Second, SWIR bands yield higher accuracy than TIR, reflecting their effectiveness in identifying cement plants via indirect indicators such as altered soil moisture and organic matter caused by dust and industrial residue. Finally, the geological SWIR ratio (Band 7:6) achieves the highest accuracy overall, indicating that spectral ratios sensitive to surface alterations offer a particularly effective means of detecting cement plant activity, complementing direct TIR-based identification of elevated kiln surface temperatures.

\subsubsection{t-Distributed Stochastic Neighbour Embedding (t-SNE)}

To better understand the feature representations extracted by the baseline CNN5 and the proposed DGCNN models, t-distributed stochastic neighbour embedding (t-SNE), introduced by Hinton and Roweis \cite{hinton2002stochastic} is applied. This statistical technique enables the visualisation of high-dimensional data by assigning each datapoint a location in a two-dimensional map, thereby revealing latent structure in the learned feature space.

Figure~\ref{fig:tsne} presents the t-SNE embeddings of feature representations learned by CNN5 (left) and DGCNN (right) for the task of cement plant detection. Each point corresponds to a sample, coloured by class label: cement (dark blue) and surrounding land cover (light blue). In CNN5, the two classes exhibit partial overlap, indicating that the learned features are less discriminative and less robust to background variation. In contrast, DGCNN yields a clearer separation between cement and land cover samples, with distinct clusters formed for each class. This suggests that embedding differential Gabor-parameterised convolutions, mixed pooling, and wider receptive fields enhances the model’s capacity to learn more structured, separable, and resilient representations for cement plant detection.

\subsubsection{Confusion Matrices}

\begin{figure}[!h]
    \centering
    \begin{subfigure}[b]{0.493\linewidth}
        \centering
        \includegraphics[width=\linewidth]{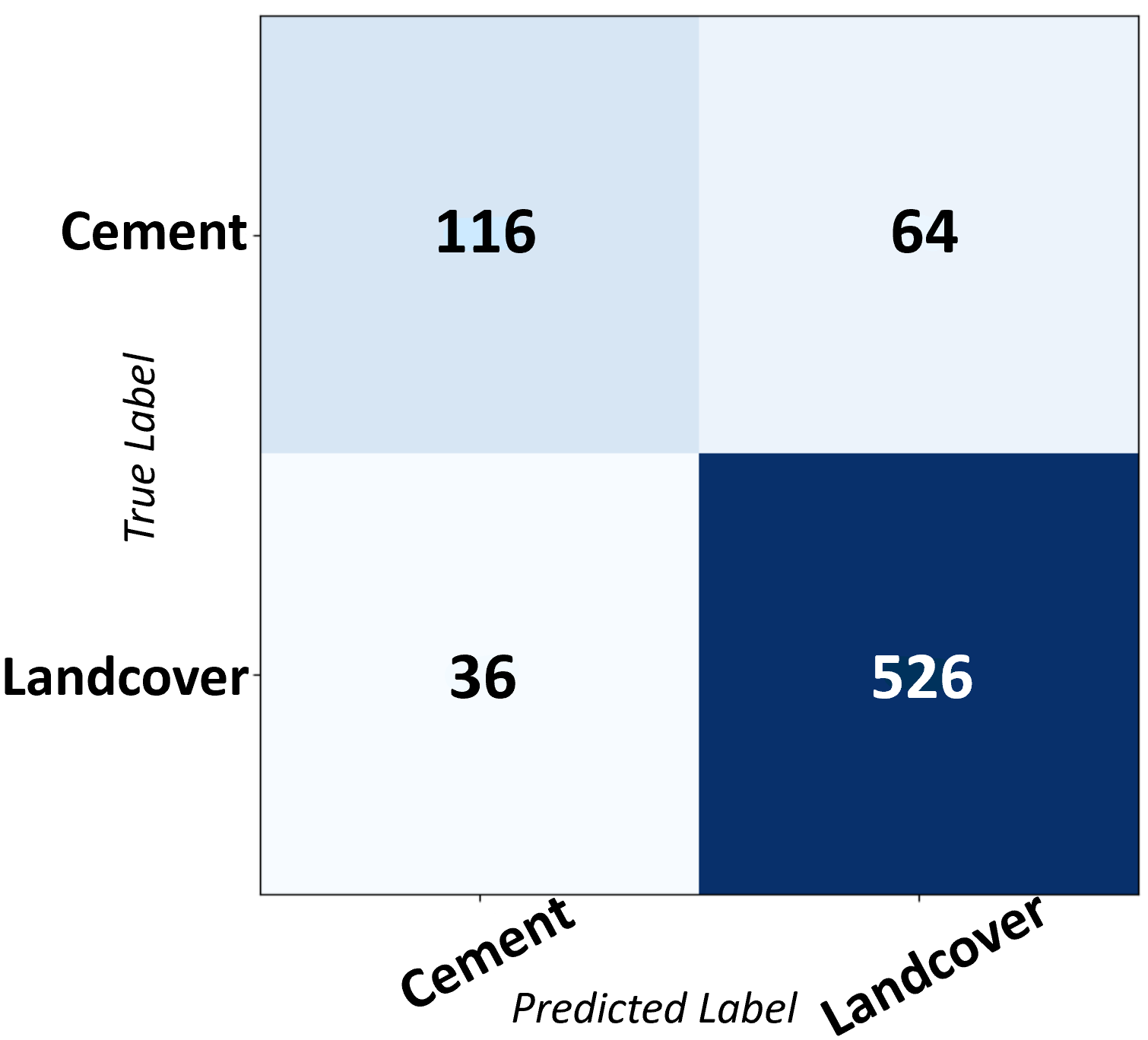}
        \caption{CNN5.}
        \label{fig:cnn5confmatrix}
    \end{subfigure}
    \hfill 
    \begin{subfigure}[b]{0.493\linewidth}
        \centering
        \includegraphics[width=\linewidth]{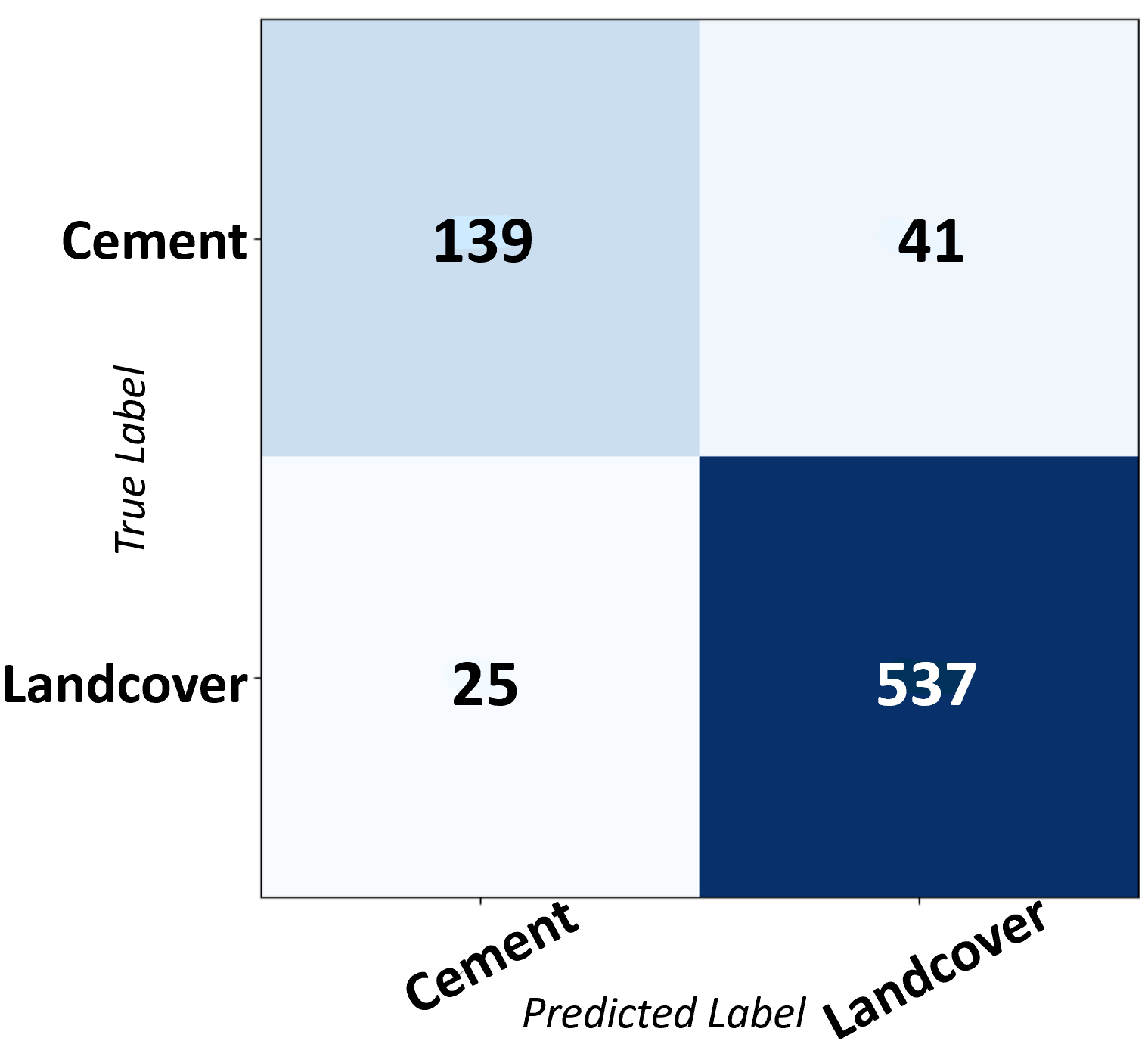}
        \caption{DGCNN.}
        \label{fig:dgcnn5confmatrix}
    \end{subfigure}
    \caption{Confusion Matrix Results a) CNN5, b) DGCNN per Cement and Landcover Classes. The Proposed DGCNN Model Outperforms the Conventional Convolutional Architecture.}
    \vspace{-0.2cm}
    \label{fig:confmatrix}
\end{figure}

To further evaluate model performance, confusion matrices of the baseline CNN5 and the proposed DGCNN models are examined, as shown in Figure~\ref{fig:confmatrix}. Each matrix reports the number of correctly and incorrectly classified samples for the cement and landcover classes.
CNN5 correctly identifies 116 cement samples but misclassifies 64 as landcover. For landcover, it achieves 526 correct predictions and 36 misclassifications. In contrast, DGCNN improves cement classification to 139 correct predictions, reducing misclassifications to 41. It also improves landcover classification, with 537 correct predictions and 25 errors.

These results highlight three key observations. First, DGCNN significantly reduces false negatives in the cement class, which is critical for reliable detection of industrial activity. Second, the model maintains high precision in landcover classification, indicating that the enhanced feature extraction does not compromise background discrimination. Third, the overall improvement in both classes supports the hypothesis that embedding differential Gabor-parameterised convolutions, mixed pooling, and wider receptive fields yields more discriminative and resilient representations.

This analysis complements the t-SNE results, confirming that DGCNN not only learns more separable features but also translates that separation into improved classification accuracy across both target and background classes

\subsubsection{Feature Extraction Analysis}

\begin{figure}[!h]
    \begin{tabularx}{\columnwidth}{
        *{3}{>{\centering\arraybackslash}X} 
    }
        \toprule
        \textbf{\small Image} &
        \textbf{\small CNN5} &
        \textbf{\small DGCNN} \\
        \midrule
        \begin{subfigure}{\linewidth}
            \raggedright
            \includegraphics[width=1.1\linewidth]{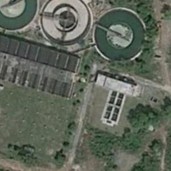}
        \end{subfigure} &
        \begin{subfigure}{\linewidth}
            \raggedright
            \includegraphics[width=1.1\linewidth]{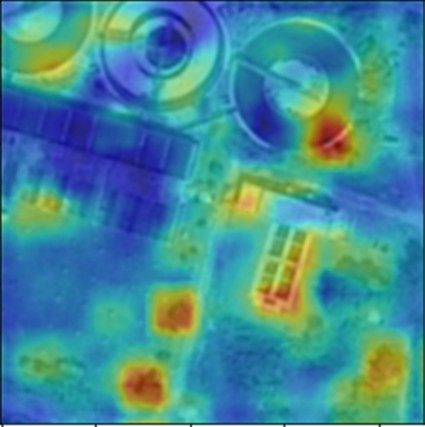}
        \end{subfigure} &
        \begin{subfigure}{\linewidth}
            \raggedright
            \includegraphics[width=1.1\linewidth]{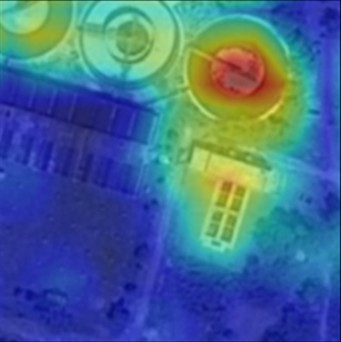}
        \end{subfigure} \\
        \begin{subfigure}{\linewidth}
            \raggedright
            \includegraphics[width=1.1\linewidth]{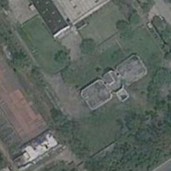}
        \end{subfigure} &
        \begin{subfigure}{\linewidth}
            \raggedright
            \includegraphics[width=1.1\linewidth]{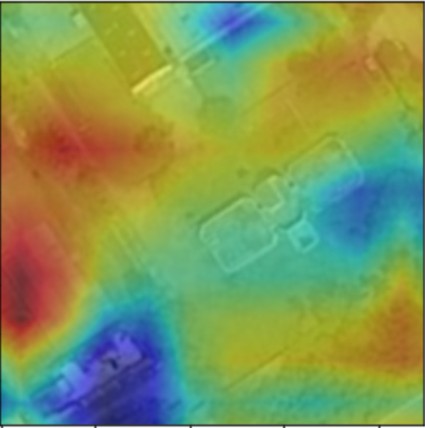}
        \end{subfigure} &
        \begin{subfigure}{\linewidth}
            \raggedright
            \includegraphics[width=1.1\linewidth]{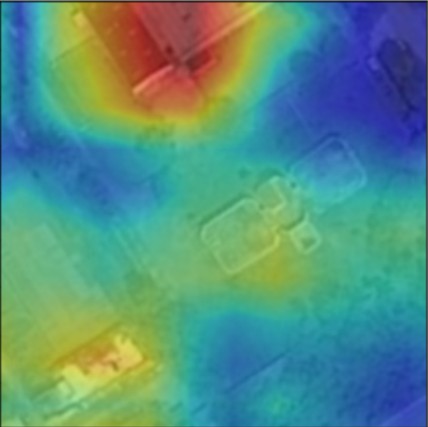}
        \end{subfigure} \\
        \begin{subfigure}{\linewidth}
            \raggedright
            \includegraphics[width=1.1\linewidth]{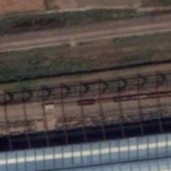}
        \end{subfigure} &
        \begin{subfigure}{\linewidth}
            \raggedright
            \includegraphics[width=1.1\linewidth]{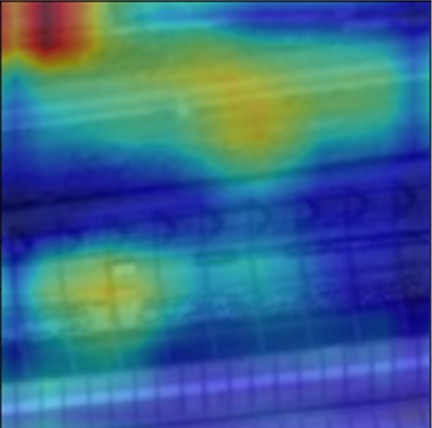}
        \end{subfigure} &
        \begin{subfigure}{\linewidth}
            \raggedright
            \includegraphics[width=1.1\linewidth]{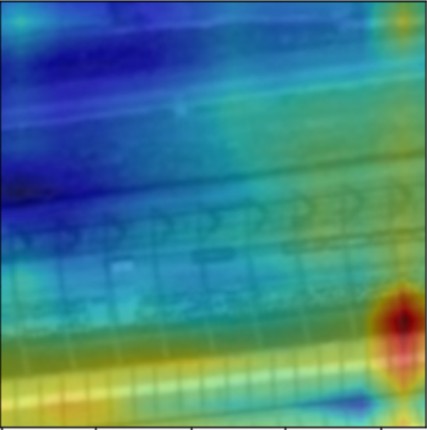}
        \end{subfigure} \\
       \bottomrule
    \end{tabularx}
    \caption{Class Activation Maps, CNN5 vs DGCNN Feature Extraction Capabilities. Proposed DGCNN: More Focused and Detailed Activation Pattern.}
    \label{fig:CAM}
\end{figure}

Class Activation Mapping (CAM, Zhou et al. \cite{zhou2016learning}) is employed to gain insight into the model's decision process by overlaying a heatmap on the original image, indicating the discriminative region used by the model when predicting that an image belongs to a particular class. Figure \ref{fig:CAM} illustrates the CAMs for images containing industrial complexes and agricultural land, comparing the feature extraction capabilities of a standard CNN with those of the proposed model. In the CNN5 column, the standard model's activation patterns are diffuse, with focus areas spread across both forested regions and agricultural lands. This scattered attention suggests inefficient feature capture, missing critical details and leading to less robust representations.

\begin{figure*}[!h]
    \centering
    \includegraphics[width=0.84\linewidth]{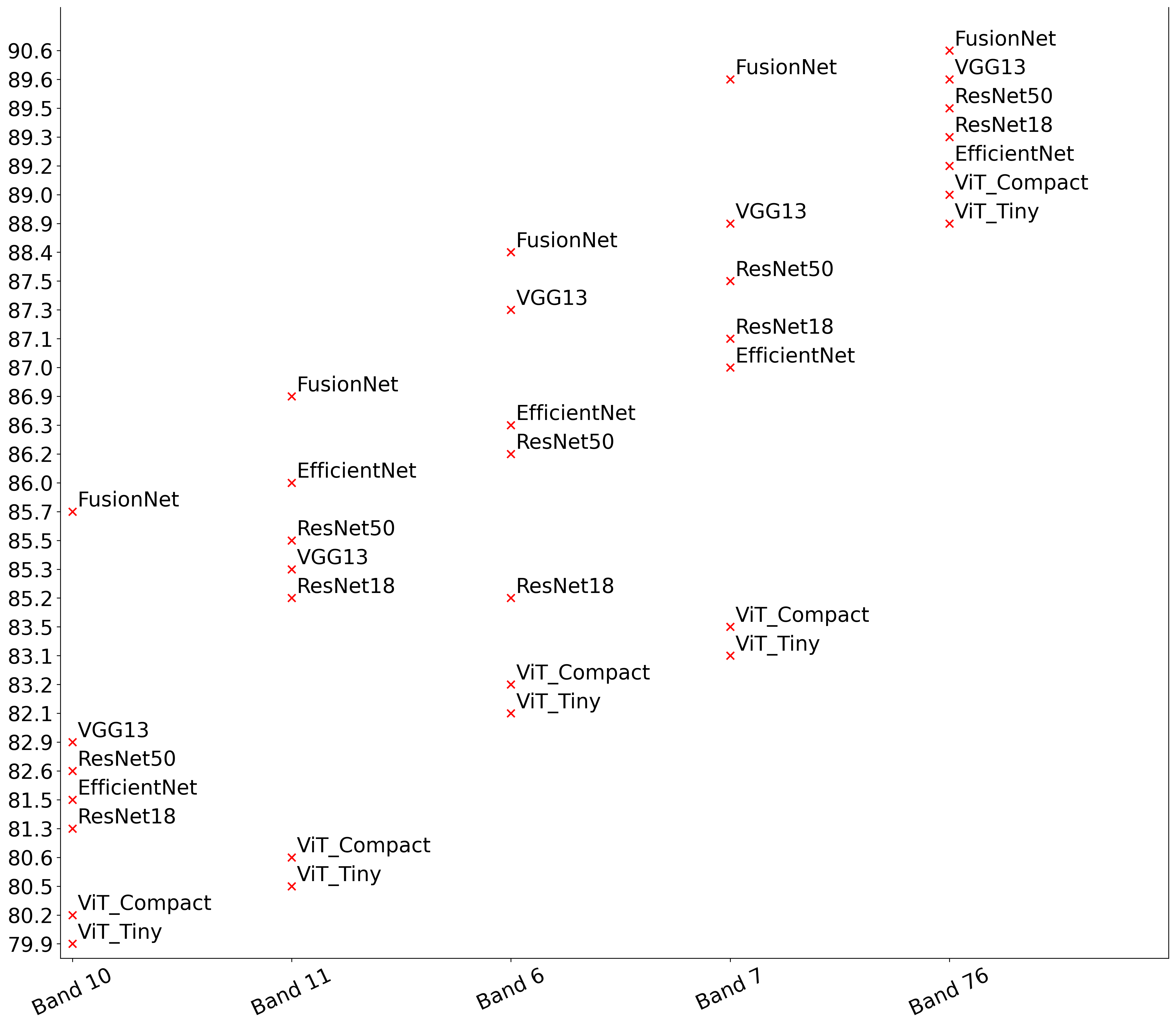}
    \caption{FusionNet Average Accuracy ($\%$) Scores Across TIR (Bands 11–10) and SWIR (Bands 7–6 and Ratio 7:6) Inputs, Evaluated per Cement and Landcover Classes. The Proposed FusionNet Outperforms all State‑Of‑The‑Art Models in Both Thermal and SWIR Datasets, with The Geological SWIR Ratio Yielding the Highest Overall Accuracy.}
    \vspace{-0.5cm}
    \label{fig:AvgAcc}
\end{figure*}

In contrast, the DGCNN model demonstrates a more focused and detailed activation pattern, particularly in areas with industrial building presence. This is expected due to the model's enhanced ability to extract salient features combined with a wider receptive field. Thus, the activation maps display more defined structures and higher contrast. This targeted focus indicates that the proposed model develops more robust and specific representations of industrial structures. The enhanced ability to concentrate on relevant features, particularly buildings, suggests better out-of-distribution generalisation, crucial for tasks like cement plant classification. Comparing these activation maps, indicates that the proposed model's architecture allows for more efficient and targeted feature extraction, potentially improving performance.

By combining all the test findings from the CAMs and t-SNE plots, it is visible that the proposed DGCNN does not just perform better, but it learns more meaningful features, enabling the model to generalise better and yield better performance in the classification task.

\subsection{State Of The Art (SOTA) Comparison}

The proposed dataset is analysed using a selection of SOTA classifiers, including convolutional architectures (VGG13 \cite{simonyan2014very}, ResNet18/50 \cite{he2016deep}, EfficientNet \cite{tan2019efficientnet}) and lightweight Vision Transformer variants (ViT\_Tiny and ViT\_Compact \cite{dosovitskiy2020image}). These models serve as baselines for evaluating the effectiveness of the proposed FusionNet architecture.

Experimental results in Figure \ref{fig:AvgAcc} demonstrate the superior performance of the proposed FusionNet architecture and the SWIR Band ratio (Band 7:6) compared to the SOTA in cement plant and landcover classification utilising Landsat 8 data. The proposed fusion model achieved maximum accuracy of $90.6\%$ utilising the proposed Band 7:6 ratio, outperforming benchmark convolutional and Transformer based architectures, including VGG13 ($89.6\%$), ResNet50 ($89.5\%$), ResNet18 ($89.3\%$), EfficientNet ($89.2\%$), ViT\_Compact ($89.1\%$), and ViT\_Tiny ($88.9\%$).

Performance analysis across different spectral bands revealed that TIR Band 10 yielded the lowest classification accuracies ($79.9-85.7\%$), whilst the novel SWIR Band 7:6 ratio demonstrated optimal discriminative capabilities with accuracy ranges of ($88.9-90.6\%$). The superior performance of SWIR compared to TIR indicates that soil moisture and organic components are more significant than temperature in distinguishing cement plants from surrounding landcover. 

Notably, the Vision Transformer baselines exhibit comparatively reduced performance on the raw spectral bands (TIR and individual SWIR inputs), which may be attributed to the limited dataset size and the weaker spatial inductive bias of attention based architectures when trained from scratch. However, on the physics-informed SWIR Band 7:6 ratio, the performance gap between convolutional and Transformer models narrows. This suggests that the proposed ratio enhances class separability at the input level, thereby reducing sensitivity to architectural bias and highlighting the value of domain-informed spectral feature engineering.

FusionNet exhibited consistent superior performance across all spectral combinations, indicating robust feature extraction capabilities. The progressive improvement in accuracy from Band 10 through the proposed Band 7:6 ratio suggests that this novel spectral combination provides the most significant features for classification. These results validate the effectiveness of both the FusionNet architecture and the SWIR Band 7:6 ratio in enhancing classification accuracy, advancing deep learning applications for industrial infrastructure mapping using multispectral satellite imagery.

\begin{table}[!h]
    \centering
    \caption{Averaged Classification Performance Results ($\%$) Over k = 5 Folds on the Proposed Geological SWIR Ratio.}
    \label{tab:AvgAcc}
    \renewcommand{\arraystretch}{1.2} 
    \scalebox{1.15}{
    \begin{tabular}{@{\hskip 9pt}l@{\hskip 9pt}c@{\hskip 9pt}c@{\hskip 9pt}c@{\hskip 9pt}c@{\hskip 9pt}} 
        \toprule
        \textbf{Models} & \textbf{Accuracy} & \textbf{Precision} & \textbf{Recall} & \textbf{F1-Score} \\
        \cmidrule{1-5}
        VGG13 & 89.57 & 84.24 & 84.30 & 84.28 \\ 
        ResNet18 & 89.32 & 84.65 & 83.92 & 84.29 \\
        ResNet50 & 89.46 & 84.55 & 82.28 & 83.40 \\
        EfficientNet & 89.18 & 84.43 & 84.48 & 84.46 \\
        ViT\_Tiny & 88.93 & 84.64 & 83.21 & 83.92 \\
        ViT\_Compact & 89.08 & 84.36 & 82.82 & 83.59 \\
        FusionNet & \textbf{90.64} & \textbf{84.83} & \textbf{86.35} & \textbf{85.59} \\
        \bottomrule
        \vspace{-0.4cm}
    \end{tabular}
    }
\end{table}

For completeness, Table~\ref{tab:AvgAcc} reports Accuracy, Precision, Recall, and F1-score (macro-averaged across both classes and averaged over five independent splits) for all models on the best performing SWIR Band 7:6 configuration. FusionNet achieves the highest performance across all four metrics, indicating consistent improvements in both overall accuracy and class balanced evaluation measures.

\begin{figure}[!h]
    \centering
    \includegraphics[width=0.45\linewidth]{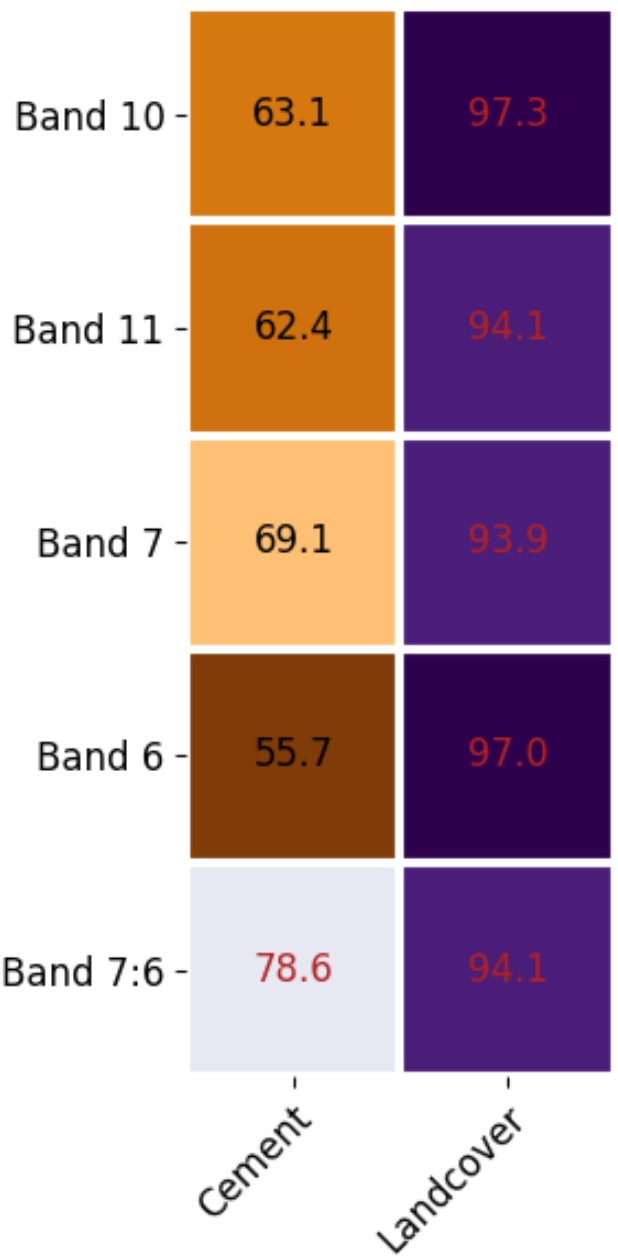}
    \caption{FusionNet Recall Scores: TIR (Bands 11-10) \& SWIR (Bands 7-6 \& Ratio 7:6) per Cement/Landcover Classes. Best Results are Obtained Using the Proposed SWIR Ratio.}
    \label{fig:recall}
\end{figure}

The recall results in Figure \ref{fig:recall} demonstrate that the proposed Band 7:6 ratio achieves the highest recall ($78.6\%$) for cement plant detection, significantly outperforming individual spectral bands. Recall measures the proportion of actual positives correctly identified, making it particularly relevant for evaluating performance on the under-represented cement class. This superior performance is followed by Band 7 ($69.1\%$), Band 10 ($63.1\%$), and Band 11 ($62.4\%$), whilst Band 6 shows the lowest recall ($55.7\%$). This validates the proposed physics-informed approach to band selection, confirming that the Band 7:6 ratio is more effective for cement plant identification.

While not the study's main focus, it is worth noting that the model demonstrates robust performance in landcover classification across all bands ($93.9-97.3\%$), indicating the general effectiveness of the proposed deep learning architecture. However, the significant improvement in cement plant detection using the Band 7:6 ratio represents the key contribution of this work, advancing the specific challenge of industrial infrastructure mapping using multispectral satellite imagery.

\subsection{Computational Complexity and Efficiency Analysis}

To assess computational overhead and deployment feasibility, Table~\ref{tab:GPUefficiency} reports model parameters, floating-point operations (GFLOPs) at $256\times256$ resolution, inference latency (ms/sample), peak GPU memory usage (MB), and training time per epoch. FLOPs were computed using \texttt{ptflops} on a representative input shape. Latency and memory were measured on an NVIDIA A100-80GB GPU using batch size 90 under identical settings. Training time per epoch was recorded from Weights\&Biases logs.

\begin{table}[h]
    \centering
    \caption{Computational Complexity and Efficiency Comparison on $256\times256$ Inputs.}
    \label{tab:GPUefficiency}
    \renewcommand{\arraystretch}{1.15}
    \setlength{\tabcolsep}{4pt}
    \begin{tabular}{lccccc}
        \toprule
        Model &
        \makecell{Params \\ (M)} &
        \makecell{FLOPs \\ (G)} &
        \makecell{Latency \\ (ms)} &
        \makecell{Peak Memory \\ (MB)} &
        \makecell{Time/Epoch \\ (s)} \\
        \midrule
        VGG13        & 160.40 & 14.86 & 0.612 & 13747 & 9  \\
        ResNet18     & 11.18  & 2.28  & 0.130 & 2565  & 7  \\
        ResNet50     & 23.52  & 5.40  & 0.422 & 10545 & 8  \\
        EfficientNet & 14.04  & 2.26  & 0.359 & 12760 & 8  \\
        ViT\_Tiny     & 0.93   & 0.36  & 0.101 & 1385  & 9  \\
        ViT\_Compact  & 0.66   & 0.29  & 0.091 & 1228  & 8  \\
        FusionNet    & 63.92  & 18.72 & 0.994 & 18860 & 48 \\
        \bottomrule
    \end{tabular}
\end{table}

FusionNet exhibits higher computational demand compared to single-stream baselines due to its five parallel sensor-specific branches and fusion module. It incurs the highest latency (0.994 ms/sample) and peak memory footprint (18.9 GB), while maintaining moderate parameter count (63.9M), lower than VGG13 (160.4M). In contrast, ResNet18 and ViT variants demonstrate substantially lower latency and memory usage, reflecting their lightweight single-branch design. EfficientNet and ResNet50 provide balanced trade-offs between computational complexity and memory consumption.

Although FusionNet introduces increased computational overhead, the observed performance gains (Section III.C) justify this trade-off for high-accuracy industrial infrastructure detection tasks. Moreover, all models can be deployed using standard deep learning frameworks on modern GPUs. Lightweight backbones operate comfortably on commodity GPUs ($\geq$8 GB), whereas FusionNet requires high-memory GPUs ($\geq$24 GB) for optimal performance.

\subsection{Impact of ImageNet-Based Transfer Learning on TIR/SWIR Performance}

Given the dataset's limited size and the strong domain differences between RGB and multispectral TIR and SWIR data, this section investigates how pretraining on the ImageNet dataset affects model performance when applied to both TIR (Band 10) and SWIR (Band 6) data. For the transfer learning (TL) experiments, all hidden layers were frozen and only the higher layers were fine-tuned using the proposed dataset.

\begin{figure}[h]
    \centering
    \includegraphics[width=1.\linewidth]{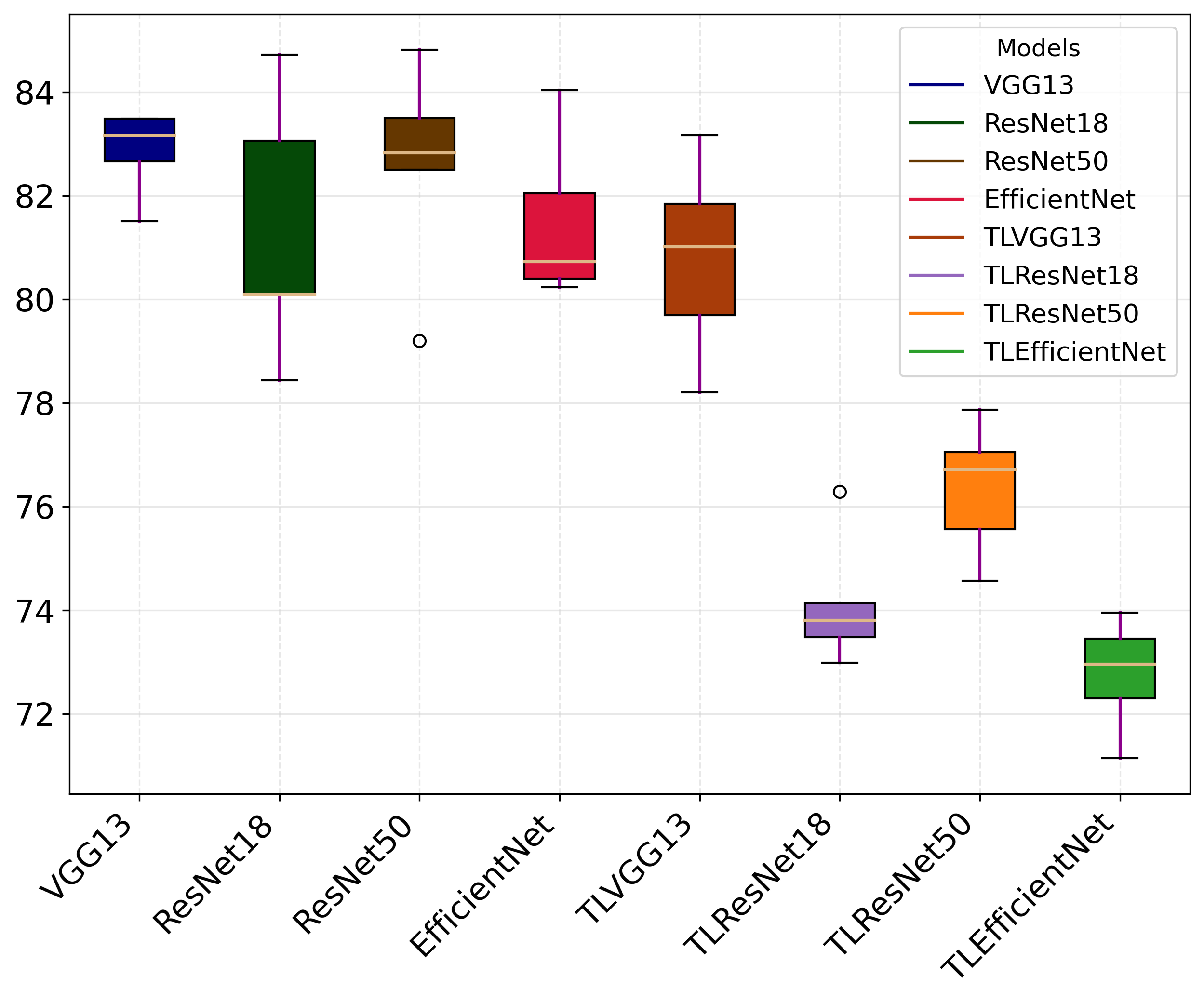}
    \caption{Comparison of SOTA CNN models (Accuracy ($\%$)) on TIR Band 10 When Trained from Scratch Versus Fine-Tuned from ImageNet-Pretrained Weights (Transfer Learning, TL). Contrary to the Typical Gains Observed in RGB Classification, TL Consistently Underperforms in This Spectral Domain Due to the Mismatch Between RGB Features and TIR Signatures.}
    \label{fig:transferlearning_B10}
\end{figure}

Figure \ref{fig:transferlearning_B10} presents the accuracies of four SOTA CNN architectures, evaluated both when trained from scratch and when initialised with ImageNet pretrained weights. The non-TL models include VGG13, ResNet18/50, and EfficientNet, while their TL counterparts (TLVGG13, TLResNet18/50, TLEfficientNet) were fine-tuned from RGB image features.

Contrary to the performance gains typically observed in RGB classification tasks, the TL variants consistently underperform relative to their non-TL equivalents. Across all architectures, the median accuracy of TL models is lower, and some also exhibit a broader interquartile range, indicating reduced optimisation stability. This systematic drop in performance is attributable to the pronounced domain gap: ImageNet pretraining encodes low- and mid-level filters optimised for three-channel RGB imagery, whereas the target TIR domain comprises thermal signatures with distinct spectral and textural properties. The pretrained filters thus bias learning towards features that are not discriminative for TIR, impeding convergence and limiting final accuracy.

\begin{figure}[h]
    \centering
    \includegraphics[width=1.\linewidth]{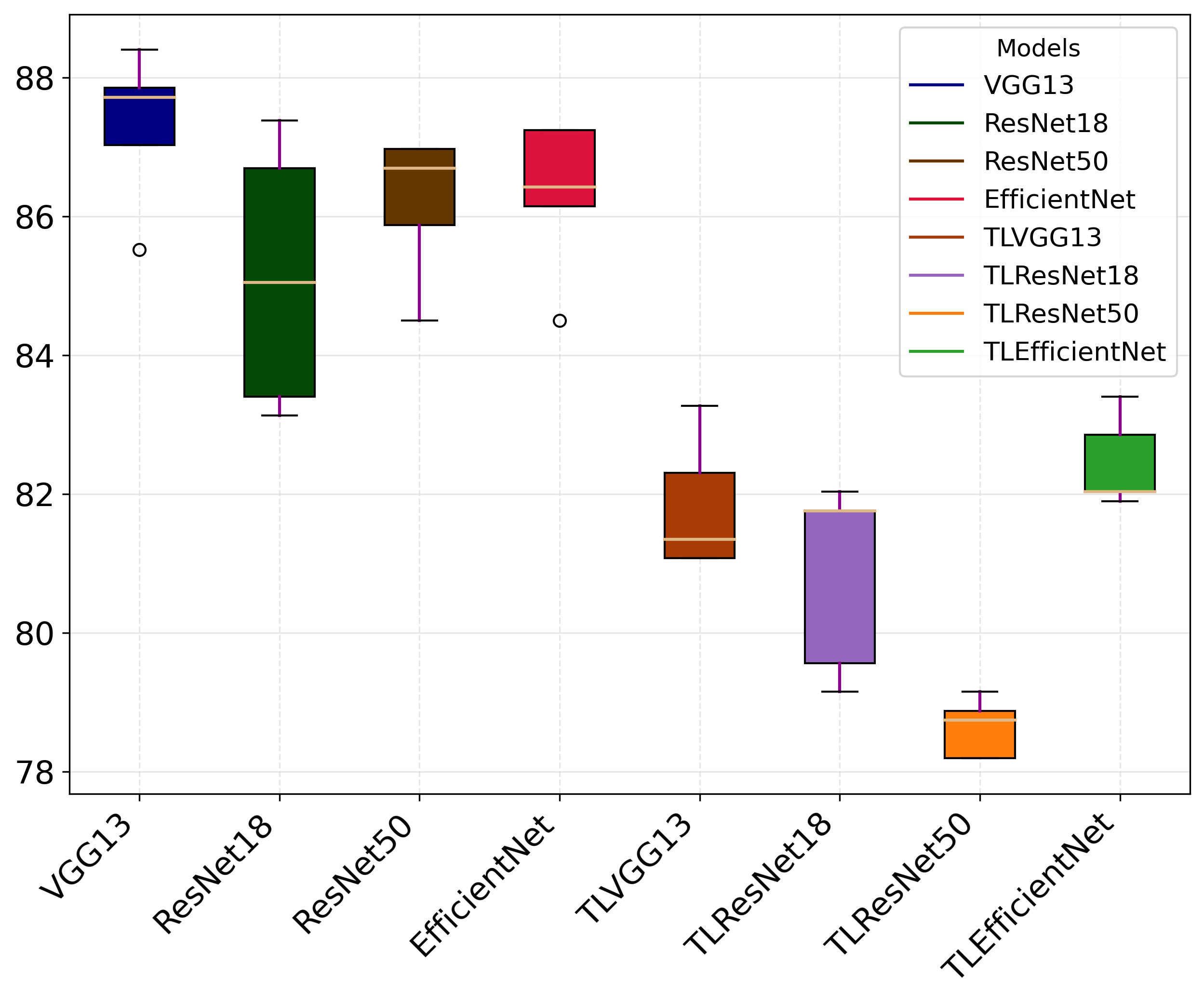}
    \caption{Comparison of SOTA CNN models (Accuracy ($\%$)) on SWIR Band 6 When Trained from Scratch Versus Fine-Tuned from ImageNet-Pretrained Weights (Transfer Learning, TL). Contrary to the Typical Gains Observed in RGB Classification, TL Consistently Underperforms in This Spectral Domain Due to the Mismatch Between RGB Features and SWIR Signatures.}
    \label{fig:transferlearning_B6}
\end{figure}

Figure \ref{fig:transferlearning_B6} shows a consistent trend for SWIR Band 6. As in the TIR case, transfer learning from ImageNet pretrained weights does not provide performance gains and consistently yields lower median accuracy across all evaluated architectures. Although SWIR captures reflectance rather than thermal emission, it still differs fundamentally from RGB imagery in spectral response and surface interaction mechanisms. Consequently, pretrained RGB filters are not optimally aligned with SWIR specific discriminative patterns, leading to reduced generalisation performance.

These findings underscore that, in domains with strong spectral and structural differences, transfer learning from RGB datasets may be detrimental. In such cases, training from scratch allows models to learn features more closely aligned with the sensing modality. This further highlights the importance of domain-aware designs, such as our physics-informed FusionNet, which directly incorporates spectral priors rather than relying on RGB-based pretraining.

\section{Discussion}

This study demonstrates that physics-informed spectral feature engineering significantly enhances deep learning performance in industrial infrastructure detection. The proposed SWIR Band 7:6 ratio consistently outperforms individual thermal and SWIR bands, indicating that indirect environmental signatures induced by prolonged industrial activity provide more stable and discriminative cues than direct thermal emissions alone. By strengthening input level separability, the ratio reduces reliance on architectural inductive biases, as evidenced by the substantially narrower CNN–Transformer performance gap on Band 7:6 compared to raw TIR and SWIR inputs.

In addition to input level structuring, the specialised DGCNN backbone, embedding differential signal-processing priors, mixed pooling, and widened receptive fields, contributes a consistent $4.1\text{--}6.8\%$ improvement over conventional CNNs, indicating that embedding differential signal-processing priors and multi-scale receptive fields enhances feature robustness in multispectral settings. FusionNet further outperforms the strongest unimodal SOTA baseline by $0.7\text{--}2.8\%$ absolute accuracy depending on spectral configuration.

The evaluation of lightweight Vision Transformer baselines further highlights the interaction between model complexity and representation quality. Under the limited data regime ($\sim4,000$ samples), Transformer based models trained from scratch underperform convolutional architectures on raw spectral bands by approximately $3.0\text{--}5.8\%$ absolute accuracy, and increasing capacity (ViT\_Tiny vs ViT\_Compact) yields only marginal gains. This contrasts with the substantially reduced CNN–Transformer gap observed on the proposed SWIR 7:6 ratio ($0.6\%$), underscoring that input level spectral structuring, rather than architectural scaling, governs performance in constrained multispectral settings.

\begin{figure}[!h]
    \centering
    \begin{subfigure}{0.32\columnwidth}
        \centering
        \includegraphics[width=\linewidth]{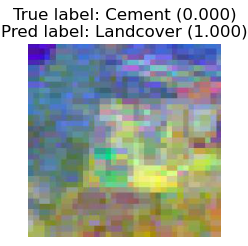}
    \end{subfigure}
    \hfill
    \begin{subfigure}{0.32\columnwidth}
        \centering
        \includegraphics[width=\linewidth]{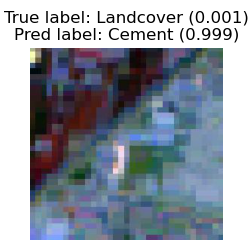}
    \end{subfigure}
    \hfill
    \begin{subfigure}{0.32\columnwidth}
        \centering
        \includegraphics[width=\linewidth]{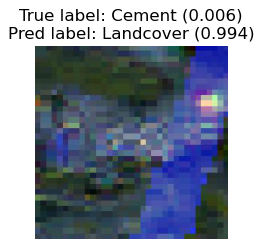}
    \end{subfigure}
    \caption{Representative Misclassified Chips With High-Confidence Errors Under Spectrally Ambiguous Conditions.}
    \label{fig:misclassified}
\end{figure}

Several limitations should be acknowledged. First, representative high-confidence misclassifications occur in spectrally and temporally ambiguous chips, as shown in Figure~\ref{fig:misclassified}. These examples exhibit heterogeneous responses, suggesting that local variability and ambiguous cement--landcover signatures remain challenging for the model. This pattern may also reflect boundary effects and overlapping spectral characteristics between the two classes.

Second, the study focuses exclusively on cement production facilities within China. Although China accounts for a substantial proportion of global cement production, environmental, geological, and climatic variability across regions may influence spectral behaviour. Future work should evaluate the robustness of the proposed SWIR ratio across diverse geographic settings.

Third, experiments were conducted using Landsat 8 imagery. While the framework is sensor-agnostic, differences in spectral response functions and signal-to-noise characteristics across sensors (e.g. Sentinel-2 or commercial high resolution platforms) may affect performance. Cross-sensor validation would strengthen the generalisability of the approach.

Finally, although this work concentrates on cement production, the underlying hypothesis, that sustained high-temperature industrial processes alter surrounding surface and soil properties, may extend to other heavy industries such as steel manufacturing, power generation, and petrochemical processing. Evaluating physics-informed SWIR ratios in these domains represents a promising avenue for future research.

\section{Conclusion}

Cement production is vital for global infrastructure but remains a major source of environmental pollution, contributing approximately 7$\%$ of global carbon emissions. Effective monitoring of these facilities is therefore essential for sustainable development. This study showed that while excess kiln heat enables detection, the soil properties captured by SWIR bands, particularly the proposed geological ratio 7:6 provide more distinctive features, boosting performance by $1\text{--}2.5\%$  over individual SWIR bands and over TIR baselines.

Systematic ablation experiments demonstrated the superior feature-extraction capability of the proposed Dilated Gabor Mixed-Pooling CNN (DGCNN) backbone. Each architectural component, embedding a differential Gabor-parameterised convolution, mixed pooling, and wider receptive field, contributed positively to accuracy, with their synergistic combination achieving consistent improvements of $4.1\text{--}6.8\%$ over conventional CNNs across spectral bands. CAM confirmed that DGCNN learns more focused and semantically meaningful features than standard CNN architectures, with activations concentrated on industrial structures rather than dispersed across irrelevant regions.

The proposed intermediate data fusion model, FusionNet, consistently outperformed state-of-the-art baselines across all five datasets. It achieved a maximum accuracy of 90.6$\%$ on the SWIR Band 7:6 configuration, representing a $1.1\%$ improvement over the strongest unimodal model, while delivering consistent $0.7\text{--}2.8\%$ gains across the remaining spectral bands. By jointly leveraging TIR and SWIR inputs, FusionNet effectively distinguished cement plants from other landcover types, offering a robust tool for environmental monitoring and sustainable infrastructure mapping.

Importantly, the reduced performance gap between convolutional and Vision Transformer models on the proposed physics-informed SWIR ratio highlights the critical role of domain-informed spectral engineering in enhancing input level separability. These findings suggest that representation design can mitigate architectural dependence, particularly in limited-data multispectral settings.

Finally, the TL investigation revealed that ImageNet pretraining consistently degraded performance on both TIR and SWIR data across multiple architectures, underscoring the domain specific nature of spectral remote sensing and the importance of training from scratch for thermal applications.

Overall, the proposed physics-aware multi-spectral fusion framework provides a robust and transferable methodology for industrial infrastructure monitoring using Earth observation data, with quantifiable accuracy gains that demonstrate both methodological and practical significance.
\small
\bibliographystyle{IEEEtranN}
\bibliography{references}

\vfill

\newcommand{\GeorgiosBioPhoto}{%
  \includegraphics[width=1in,height=1.25in,keepaspectratio]{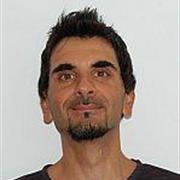}%
}

\begin{IEEEbiography}[\GeorgiosBioPhoto]{Georgios Voulgaris}
is a Postdoctoral Researcher in Computer Vision and Deep Learning at the University of Oxford, Oxford, U.K. He received the Ph.D. degree in Informatics from the University of Sussex, Brighton, U.K., in 2023. His doctoral research focused on developing deep learning models that extract transferable feature representations for domain adaptation and semantic segmentation. In 2022, he held a DISCnet Research Internship at the Satellite Applications Catapult, where he developed deep learning models for industrial remote sensing using multispectral and thermal Earth observation data. 

His current research investigates how inductive biases from visual perception and physical signal formation shape representation learning in deep neural networks, with a focus on perceptually and physically grounded architectures to improve robustness, interpretability, and generalisation under real-world distribution shifts.

 Dr. Voulgaris serves as a Technical Committee Member for CVPR EarthVision and as Co-Organiser of the ICIP Workshop on Computer Vision for Ecological and Biodiversity Monitoring. He is a recipient of the EPSRC Doctoral Training Studentship, the DISCnet Doctoral Training Scholarship, and the Best Research Proposal Award at the Machine Learning Summer School, MLSS-Indonesia 2020.
\end{IEEEbiography}

\end{document}